\documentclass[lettersize,journal]{IEEEtran}
\usepackage{amsmath,amsfonts}
\usepackage{algorithmic}
\usepackage{algorithm}
\usepackage{array}
\usepackage[caption=false,font=normalsize,labelfont=sf,textfont=sf]{subfig}
\usepackage{textcomp}
\usepackage{stfloats}
\usepackage{url}
\usepackage{verbatim}
\usepackage{graphicx}
\usepackage{cite}

\usepackage[T1]{fontenc}
\usepackage{subfloat}
\usepackage{booktabs}
\usepackage{multirow}
\usepackage{bbding}

\usepackage{algorithm}
\usepackage{algorithmic}

\usepackage{tikz}
\usetikzlibrary{shapes,arrows}

\begin{document}

\title{Multitask Learning for SAR Ship Detection with\\ Gaussian-Mask Joint Segmentation}

\author{\IEEEauthorblockN{Ming Zhao, Xin Zhang, Andr\'{e} Kaup, \emph{Fellow, IEEE}}
	
\thanks{This work was supported in part by the National Natural Science Foundation of China under Grant 62271302, Grant 62101316; in part by the Shanghai Municipal Natural Science Foundation under Grant 20ZR1423500. (Corresponding author: Ming Zhao)}
\thanks{Ming Zhao and Xin Zhang are with the Department of Information Engineering, Shanghai Maritime University, Shanghai 201306, China. (e-mail: zm\_cynthia@163.com).}

\thanks{A. Kaup is with the Chair of Multimedia Communications and Signal Processing, Friedrich-Alexander University Erlangen-N$\ddot{u}$rnberg,
	Cauerstr. 7, 91058 Erlangen, Germany. }
\thanks{Manuscript received April 19, 2021; revised August 16, 2021.}
}

\markboth{Journal of \LaTeX\ Class Files,~Vol.~14, No.~8, August~2021}%
{Shell \MakeLowercase{\textit{et al.}}: A Sample Article Using IEEEtran.cls for IEEE Journals}


\maketitle

\begin{abstract}
Detecting ships from synthetic aperture radar (SAR) images is inherently subject to the limitations of SAR's imaging mechanism. SAR object detection technology has rapidly advanced in recent years due to deep learning based techniques for detecting objects from optical images.
SAR ship detection still faces some challenges due to the strong speckle noise, complex surroundings, and variety of scales.
This paper proposes a multitask learning framework for object detection (MLDet)  detect ships in SAR images.
The proposed end-to-end framework  consists of object detection task, speckle supression task and target segmentation task. 
Firstly, an angle classification loss with aspect ratio weighting is explored during object detection  to  improve the accuracy by making the detector sensitive to the periodicity of angular and the aspect ratio of objects.
Secondly, the speckle supression task employs a dual-feature fusion attention mechanism to suppress noisy background information and fuse shallow features and denoising features,
which helps MLDet be more robust to speckle noise. 
Thirdly, the target segmentation task with rotated Gaussian-mask is explored to further help the detection network to extract the regions of intersect from the cluttered background, as well as improving the detection efficiency through pixel-by-pixel prediction. 
The rotated Gaussian-mask for ship modeling ensures that the center of  a ship  has the highest probablities to be labeled as an object, and the probablities of the remaining regions are gradually reduced under a Gaussian distribution.
Assisted by these two subtasks, the shallow level features are robust to speckle noise and reliably support deep level feature learning.
In addition, the weighted rotated boxes fusion (WRBF) strategy  is adopted to combine the predictions of multi-direction anchors for rotated objects, and eliminate the anchors beyond the boundary as well as the anchors with high overlap rates but low scores. 
A large number of experiments and comprehensive evaluations
on SAR ship detection datasets SSDD+ and HRSID have shown the effectiveness and superiority of the proposed method. 
The code is available from \emph{https://github.com/zx152/MLDet} .
\end{abstract}


\section{INTRODUCTION}
\IEEEPARstart{S}{YNTHETIC} aperture radars (SAR) can be used in all weather conditions, both day and night, and are capable of producing high-resolution images. 
Maritime surveillance, ocean monitoring, and fishery control have benefited greatly from the rapid development of spaceborne SAR systems, e.g., RADARSA T-2, TerraSAR-X, Sentinel-1, and Gaofen-3.  However, the special imaging mechanism inherent in SAR imagery may make it difficult to identify targets of interest in massive SAR images. Searching for targets by eye would be time-consuming and impractical in such scenarios. 
It has been the focus of research to discover and identify various targets in SAR images as quickly and accurately as possible. Specifically, SAR ship detection has become increasingly popular due to its practical applications, including marine monitoring, maritime management, and intelligence gathering \cite{Brusch2011Ship1}.
Therefore, Ship detection in SAR imagery has gained increasing attention in recent years, and more and more researches have been studies \cite{2018Detection3}.
\par
As a general rule, the traditional methods of ship detection rely primarily upon a statistical analysis of the image pixels, which can be mainly divided into two groups \cite{2003Survey4} $-$ \cite{Chan2008Automatic6}. 
The first group is the threshold-based methods, which firstly model the background clutter by the filtering theory of constant false alarm rate (CFAR), and then adaptively adjust the detection threshold to distinguish the ship objects from the backgrounds \cite{2018An7}, \cite{Jo2010A8}.
These threshold-based methods perform well in offshore areas in SAR images, but easily fail to distinguish between coast and ship in inshore areas.
Besides that, it is difficult to estimate parameters because of the higher complexity and higher fitting accuracy for the modeling process. 
The accuracy of background clutter modeling is often weighed against the complexity of the computation by many researchers\cite{2016Statistical10}, \cite{Gao2017Scheme11}. 
Li and Zelnio \cite{2002Target12} proposed a SAR object detection method based on the generalized-likelihood ratio test, which statistically models both of the background clutter and ship targets.
As well as threshold-based methods, there is another group methods based on geometric characteristics of the targets and backgrounds, such as standard deviations and non circularity\cite{2017Noncircularity13}, \cite{2019Imbalanced14}.  
Principal component analysis \cite{2017Ship15} and Bayesian theory \cite{2010Ship16} are often employed in these methods to extract various statistical characteristics, i.e., shape, area, size, and texture. After that, the strategy of template matching is adopted to conduct the SAR ship detections. 
The statistical characteristics of SAR images by these  traditional methods are fully exploited, and generally produce good performance for some spescific scenes, particularly asisted with prior information. 
However,  their adaptability of manual-designed features is insufficient, and cannot guarantee robust performance for complex scenarios. 
Besides that, traditional methods usually need  complicated steps, and consequently can not directly be applied to the real-time detections.

\par In recent years, there has been a significant amount of attention paid to the target detections for remote sensing images based on convolutional neural networks (CNNs).
In comparison with traditional methods, CNN is capable of learning the  features  from large amounts of image data and  determining the target locations with greater accuracy and robustness.
In general, CNN-based detection methods can be divided into two groups based on the detection strategy, i.e., one-stage methods and two-stage methods.
The two-stage methods firstly generate amounts of proposals to be classified and regressed.
R-CNN \cite{2013Rich8} is the first two-stage method to be applied to the field of object detection.
As an extension of R-CNN \cite{2013Rich8} and SPPNet \cite{Kaiming2015Spatial16}, Fast R-CNN \cite{2015Fast9} proposes a region of interest (RoI) pooling layer to enhance the speed and accuracy of detection. 
To extract region proposals, Faster R-CNN \cite{2017Faster10} replaces the selective search by the proposed region proposal network (RPN). 
Based on Faster R-CNN, Mask R-CNN  \cite{2017Mask17} adds a mask branch to predict an object mask. 
The use of region-based fully convolutional networks (R-FCNs) \cite{2016R18} has been shown the ability of significantly accelerating the detection speed through sharing calculations derived from Faster R-CNN. 
In addition, the feature pyramid network (FPN) \cite{2017Feature19} also addresses the problems of multiscale changes in object detection. 
Aiming at solving this problem, this network integrates multilayer feature information that widely adopted by other  algorithms.
Libra R-CNN \cite{2020Libra20} overcomes the imbalanced problem during training by using IoU balanced sampling, balanced feature pyramids, as well as the balanced first loss. 
By utilizing a multistage architecture, Cascade R-CNN  \cite{2017Cascade21} avoids the problems associated with overfitting during the training stage as well as the quality mismatch during the inference stage. 
Besides that, it has become increasingly necessary to improve deep learning models that can be deployed in real-time scenarios or scenarios using as few resources as possible.
Ghosh et al.\cite{Deep network} examined pruning to reduce the number of calculations and required weights by employing deep learning models for object detection.  Quast et al. \cite{2010Real} further accelerated the object detection process by utilizing parallel computing.
Group Lasso\cite{Yuan2006ModelSA} is an efficient regularization to learn sparse structures. Liu et al. \cite{Liu2015SparseCN} utilized group Lasso to constrain the structure scale of LRA. To adapt DNN structure to
different databases, Feng et al. \cite{Feng_2015_ICCV} learned the appropriate number of filters in DNN.

\par Even though two-stage methods are capable of obtaining accurate detection results for remote sensing images, the high time complexity makes them unsuitable for most realistic applications.
The one-stage methods are different from  two-stage methods in that the detection result is obtained directly from the input image instead of generating candidate regions. 
For instance, YOLO series algorithms \cite{2016You12}–\cite{2018YOLOv314}, \cite{2020YOLOv422} treat detection as a regression task and employ CNNs to achive effective detections. 
The single-shot multibox detector (SSD) \cite{2015SSD11}  detects objects of different sizes  assisted with the feature maps of different scales.
As an one-stage algorithm,  RetinaNet  \cite{2017Focal15} alleviates  the imbalance problem  between the positive samples and the negative samples by applying the focal loss. 
In recent years, anchor-free models have come out, which has no more need the designed anchors according to the prior knowledge. 
The key-point based method CornerNet\cite{2020CornerNet23} and the anchor-point-based methods, e.g., FCOS \cite{2020FCOS24} and CenterNet \cite{2019Objects25} are these representative anchor-free algorithm.

\par
\begin{figure}[th]
	\centering
	\subfloat[]{
		\includegraphics[width=0.49\linewidth]{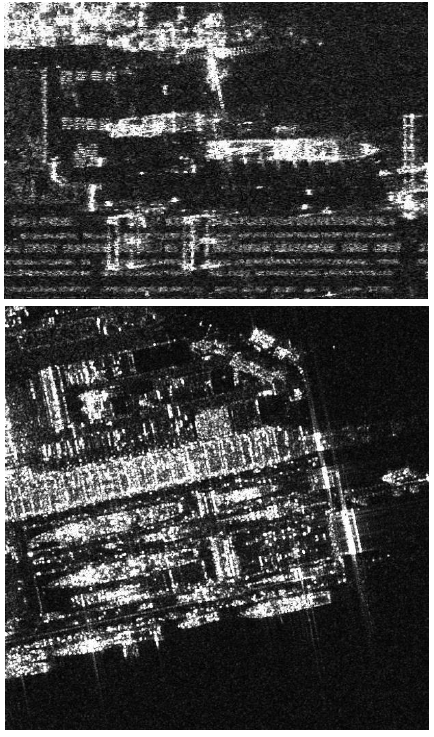}
		\label{fig:dataset problem a}
	}
	\subfloat[]{
		\includegraphics[width=0.49\linewidth]{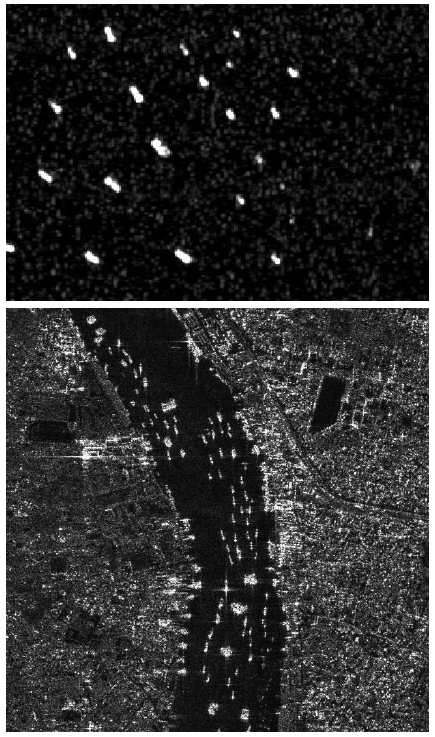}
		\label{fig:dataset problem b}
	}
	\caption{Examples of SAR images. (a) SAR images with complex backgrounds. (b) SAR images with multiscale ship targets.}
	\label{fig:dataset problem}
\end{figure}

 Even though the CNN-based detection algorithms perform better than the  above traditional methods  \cite{2003Survey4}-\cite{2010Ship16}, they cannot be directly applied to SAR ship detection.
Firstly, owing to the unique nature of SAR imaging technology, there are complex backgrounds with considerable noise in SAR images, particularly in the shore areas, as shown in Fig. \ref{fig:dataset problem}(a). 
Moreover, ship detection in SAR images will produce unreliable false alarms due to the presence of a wide range of disturbances, including sea clutter, islands, and land. 
Secondly, as shown in  Fig. \ref{fig:dataset problem}(b), due to the multi-resolution imaging modes and different shapes of ships, SAR images often contain multiscale targets. 
Upon mapping small targets to the final feature map, little information is available to refine the location and classify the objects, which causes high false negatives and low true positives. 
Thirdly, the strong speckle noise presented in SAR images hinders the learning of low-level semantic features for the detection of objects at a higher level, as well as affecting low-level feature extracting in shallow layers. 
For instance, both ships inshore and buildings on the shore  appear as the bright white speckles as if they are floating in the water. Therefore, the speckle noise will easily destroy the object boundaries, which renders them difficult to learn.

\vspace{-0.5cm}

\par Aiming to solve the problems mentioned above, researchers attempted to improve the performance for CNN-based SAR ship detection. 
An et al. \cite{2018Ship2}  firstly introduced  CNNs to segment land and sea, and then applied CFAR to detect ship, which improved its performance at sea-land junctions.
Faster R-CNN combined with CFAR was improved by Miao et al.\cite{2017A35}. 
As a result of the improved Faster R-CNN, small targets can now be detected more effectively in the protection window of CFAR. Increasingly, researchers are focusing on end-to-end  deep learning based methods for SAR ship detection, which gets benefit from the good development for CNNs and the rapid enrichment of SAR image datasets. 
According to Wang et al. \cite{2019A}, a two-stage detector extracts foreground proposals and distinguishes objects from virtual shadows via coarse and fine recognition stages. 
After introducing saliency information into the feature extraction, the network proposed by  Duet al. \cite{2020Saliency} was able to focus on the object areas more effectively. 
Zhao et al. \cite{2020Attention} developed an attention receptive block applied to SAR ship detection, and a fine-grained feature pyramid  was integrated into the attention receptive block for feature extraction. 
Besides that, more and more single-stage detectors are applied to SAR ship detection in place of two-stage detectors on the advantage of the faster speed.
A SAR ship detector proposed by Deng et al. \cite{2019Learning} learned from scratch, and used positional score maps to encode the positional information into ship proposals. 
Chen et al. \cite{2020SAR} included a deconvolution module as well as a prediction module based on SSD. 
Yang et al. \cite{2021A}  proposed R-RetinaNet as a solution to the problems of the feature scale mismatches, conflicting  subtasks of learning, as well as the unbalanced samples of positive responses. 
Chen et al. \cite{2020Learning} explored a  novel attention mechanism to improve the performance of detectors, which combines knowledge distillation with attention mechanisms.

\par 

According to the related works mentioned above, it is clear that
the complex background, speckle noise, and the diversity of ship scales impact on the feature learning of objects in SAR images.
Treating SAR ship detection as a single problem will ignore the close connections between  denoising of speckles, target segmentation, and the detection of objects.
In this paper, motivated by the observation that the cooperative evolution between the low-level features and the high-level features contributes to feedback-driven feature learning with the end-to-end frameworks, a multitask learning ship detection (MLDet) framework is proposed. It  consists of three subtasks, i.e., an object detection module, a denoised feature fusion module, and a target segmentation module. 
The contributions of this work are as follows:

\begin{enumerate}
\item  A novel multitask learning framework for SAR ship detection, assisted with speckle supression task and target segmentation is presented to solve the object detection problem in SAR images. Compared to traditional single-task learning, the multitask learning has more powerful ability of learning features for  ship detection in SAR images.  

\item An angle classification loss with aspect ratio weighting is explored to improve the robustness of detections, especially for SAR ships with the small aspect ratios. This further improves the accuracy on the aspect of making the detector more responsive to the periodicity of angular angles and aspect ratios.

\item  The denoised feature fusion (DFF) module is proposed to reduce the interference of complex backgrounds and enhance the salient features of target ships. 
A dual-feature fusion attention mechanism is designed into the DFF module to fuse the shallow features and the denoised features, which effectively makes the networks pay more attention to the feature extraction for targets. 	
	
\item A target segmentation module with rotated Gaussian-mask is explored to furtherly help the detection network to extract and propose the regions of intersect from the cluttered background. Using a rotated Gaussian mask for ship modeling ensures that the ship center is most likely to be labeled as an object. The probabilities of the remaining regions are gradually reduced according to the  distribution of Gaussian.

\item The commonly used fusion strategy non maximum suppression (NMS) places too much emphasis on classification confidence, but pays insufficient attention to localization accuracy.
In order to improve this, the weighted rotated boxes fusion (WRBF) strategy  is adopted to combine the predictions of rotated object detection and target segmentation, expecting to increase the generalization ability of the SAR ship detector.	
	
\end{enumerate}

\par This paper is organized as follows: Section II introduces the proposed method in detail, Section III provides the experimental results and the analysis, and Section IV draws the conclusions.

\begin{figure*}[th]
	\centering
	\includegraphics[width=1.02\linewidth]{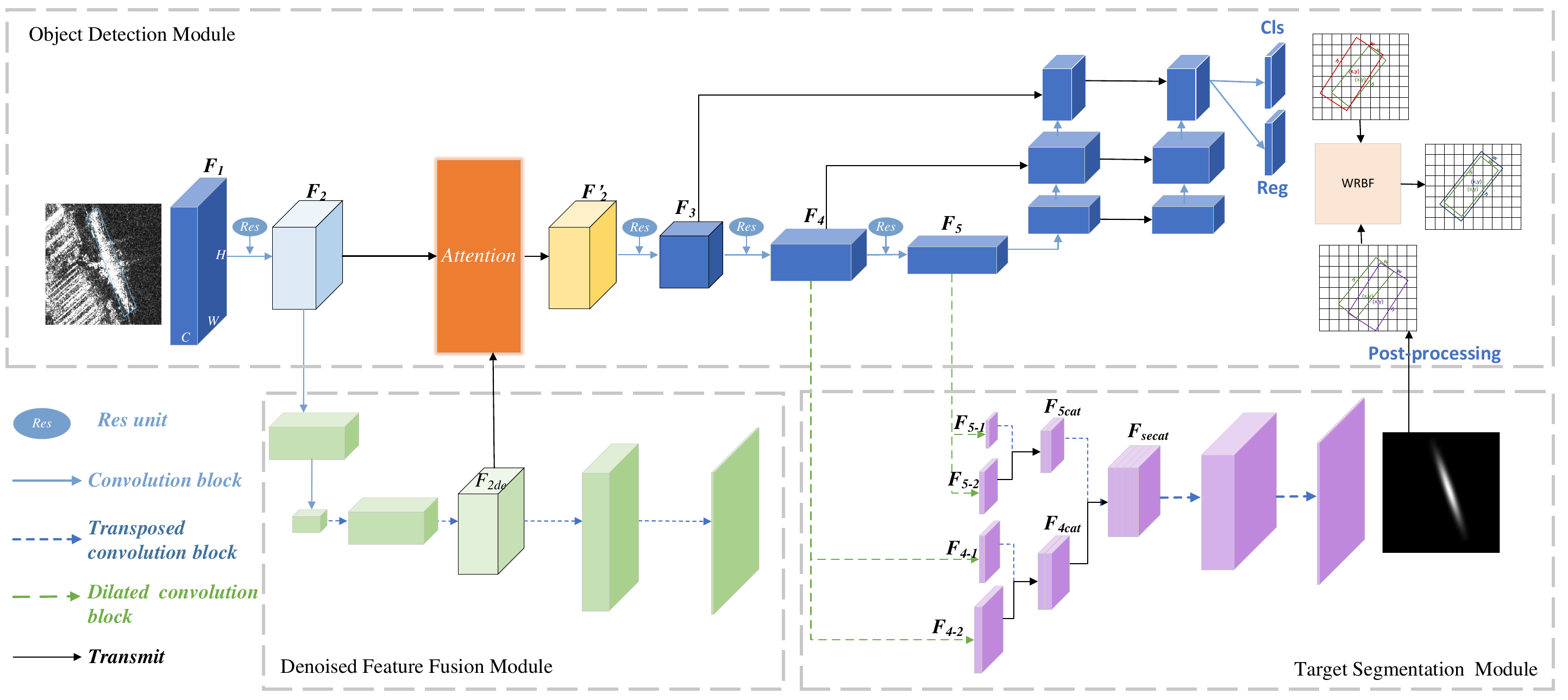}
	\caption{The architecture of the proposed MLDet including, object detection module,  denoised feature fusion module, and target segmentation module.}
	\label{fig:backbone}
\end{figure*}

\section{METHODOLOGY}
In order to realize the effective SAR ship detection  with cluttered backgrounds, speckle noise and a diversity of ship scales, a multi-task learning detection framework called MLDet is constructed in this paper. MLDet accomplishes this goal by jointly learning  object detection task, speckle suppression task, and target segmentation task. As shown in  Fig. \ref{fig:boxes}, the proposed MLDet consists of three modules, i.e., object detection module, denoised  feature fusion module and target segmentation module.

\subsection{Object Detection Module}
The features of the orginal image are extracted in the object detection module, which provide important information for simultaneously learning speckle suppression, target segmentation and object detection. 
This module consists of CSPDarknet and  a common block shared with the other two modules. 
The  proposed network structure is preserved by equipping the common block with residual blocks of feature extractor, instead of designing them separately. 
In particular, twenty-three residual blocks are used in five residual stages denoted as $F_1$, $F_2$, $F_3$, $F_4$ and $F_5$ during  feature extraction. 
The shallow features may contain more spatial information conducive to speckle suppression, while the deep features extracted in deep layers contain less spatial information. Therefore, we choose the first eleven convolutional layers in the object detection module to compose the common block and make the feature map $F_2$ as the output of this module, as shown in Fig. \ref{fig:backbone}.
The resulting feature map $F_2$ is simultaneously delivered to the DFF module for speckle suppression, and then delivered to the dual-feature fusion attention mechanism (DFA) for the fusion of two feature layers. The fused feature map $F_{2cat}$ is then sent to the remaining residual blocks for remained feature extraction.

\par
\begin{figure}[H]
	\centering
	\includegraphics[width=0.65\linewidth]{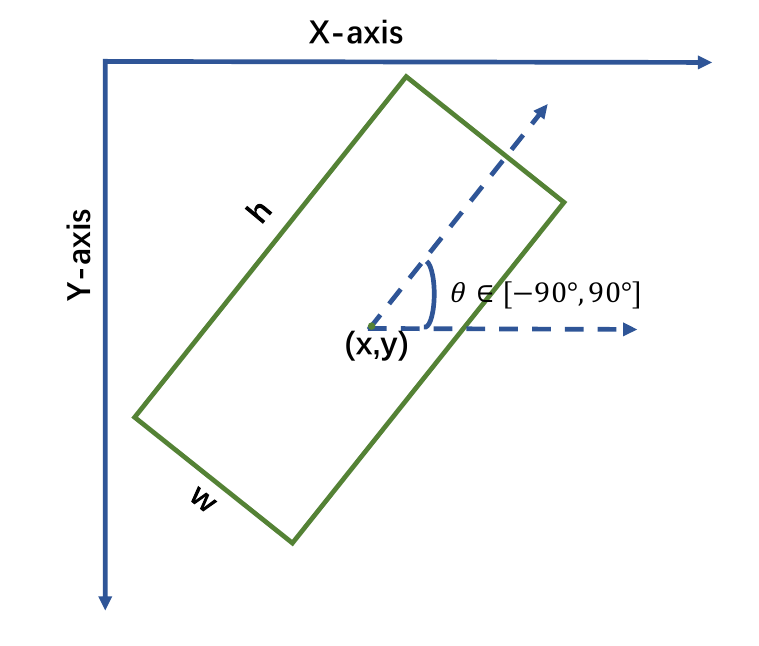}
	\caption{The definition of rotated rectangle $\theta$ .}
	\label{fig:boxes}
\end{figure}

  Following the acquisition of the feature map, the prior anchors are produced  at a place corresponding to the anchor point associated with each pixel in the feature map. 
The intervals of the anchor points are dependent on the spatial connections between the original image and the corresponding feature map, which is well explained and verified in the previous researches.
In most previous studies, anchors based on horizontal bounding boxes (HBB) were used to detect ship targets. 
However, recent researches have demonstrated that oriented bounding boxes (OBB) can describe targets more accurately by adding a new orientation angle parameter. 
It is also important to note that the OBB can also be effective in improving detection performance, due to the good adaptations to densely arranged ships.
The OBB for object detection shown in Fig. \ref{fig:boxes} is defined by the following parameters, i.e., the coordinates of  bounding box's central point (${x,y}$), the bounding box's height and width ${h}$ and  ${w}$, as well as the rotation angle $\theta$. 
As defined by the rotation angle $\theta$ of the OBB, it is the angle between  the horizontal axis and the long axis of the OBB that ranged from  $-90^{\circ}$ to $90^{\circ}$. 

\begin{figure}[th]
	\centering
	\subfloat[]{
		\includegraphics[width=0.45\linewidth]{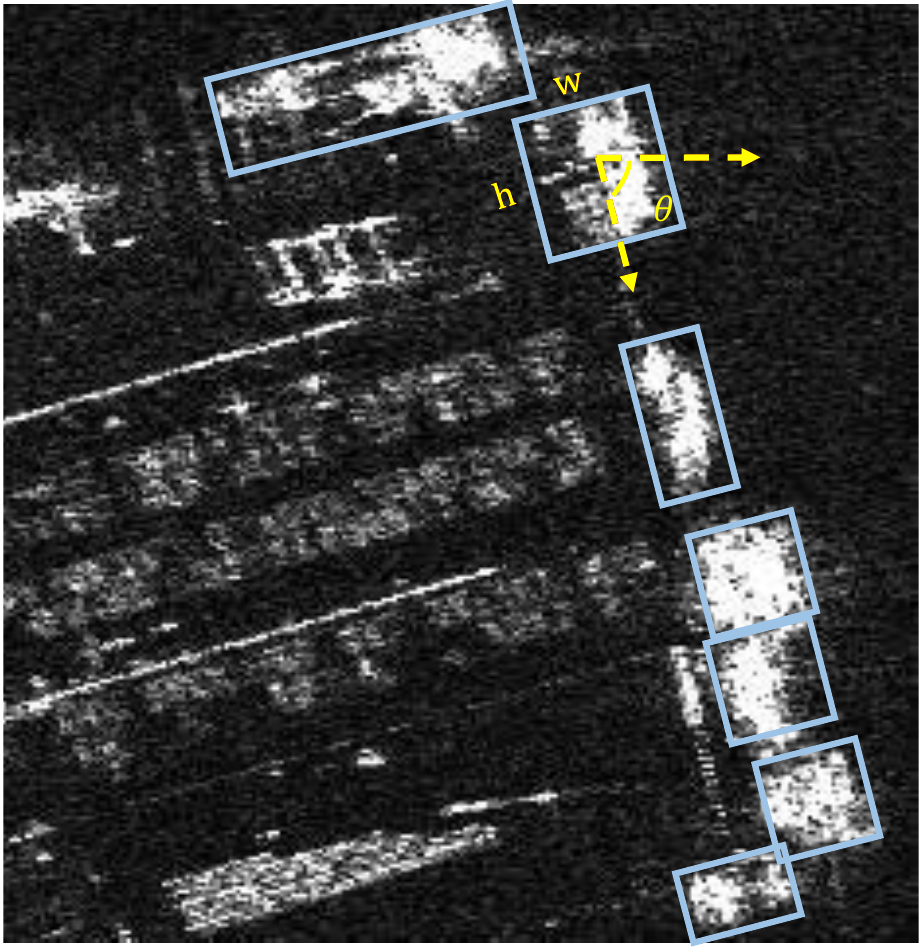}
		\label{fig:data problem a}
	}
	\subfloat[]{
		\includegraphics[width=0.45\linewidth]{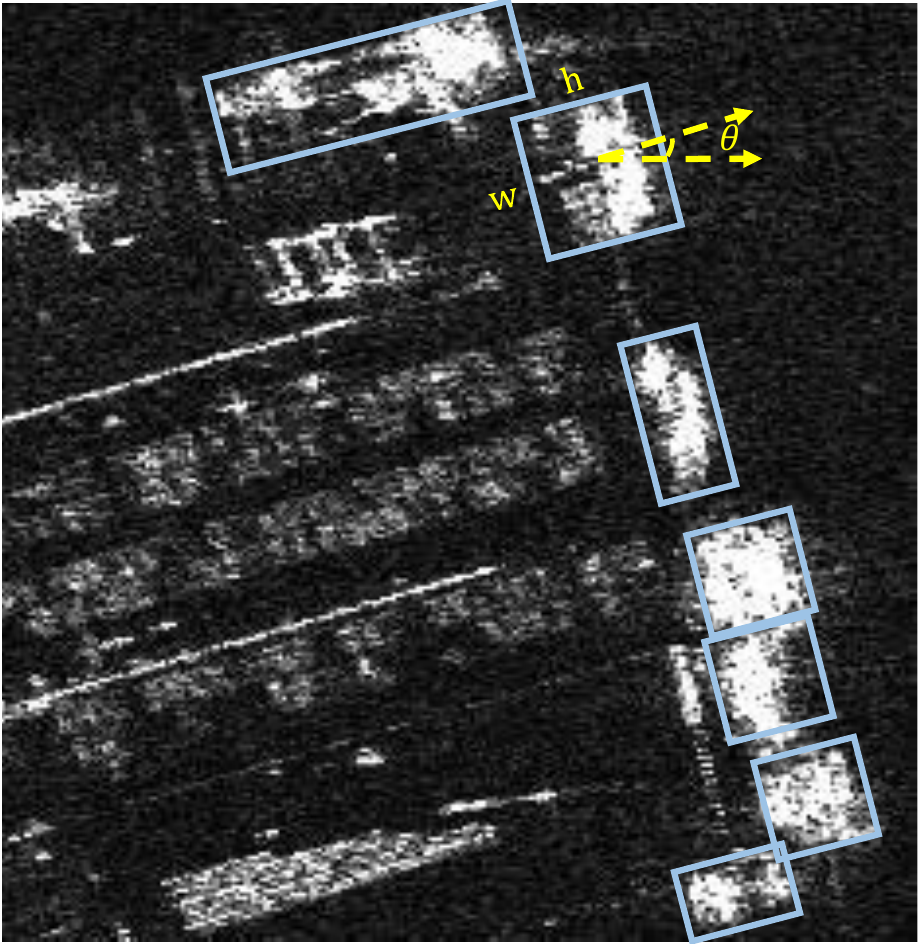}
		\label{fig:data problem b}
	}
	\caption{A demonstration of the limitations of the long-edge definition method for square-like boxes, which have a high IoU but a large training loss due to their angle mismatch. (a) Ground truth. (b) Prediction boxes.}
	\label{fig:data problem}
\end{figure}

Although this definition method avoids the exchangeability of edges problem (EoE problem), it is not suitable for square-like boxes demonstrated in Fig. \ref{fig:data problem}.
Fig. \ref{fig:data problem}(a) and (b) are the ground truth and candidate prediction bounding box respectively, while the correponding yellow labeled square-like boxes have the same height ${h}$ as well as width ${w}$. 
Both of the aspect ratio between square-like boxes are close to 1, but their angles are $-70.6^{\circ}$ and $19.7^{\circ}$ respectively. 
As a result of calculating Intersection-over-Union (IoU) and regression or classification loss, the training produces relatively high loss values by these square-like boxes, although the IoU between  the predicted bounding box and the ground truth is close to 1.
Thus, the prediction bounding box shown in Fig. \ref{fig:data problem}(b) increases the difficulty when it comes to predict objects with small aspect ratios.

In this paper, we propose to solve this problem by adding a periodic trigonometric function to the long-side definition method in order to define square-like boxes. As the aspect ratio increases, this phenomenon becomes less noticeable. We define an angle classification loss with aspect ratio weighting (ARW) as follows: 
\begin{equation}
	\begin{aligned}
		\mathcal{L}_{\theta}\left(\theta,\hat{\theta} \right) & = \left|\sin \left(\alpha\left(\theta - \hat{\theta}\right)\right)\right|\times {\mathcal{L}_{smoothl1} \left(\theta,\hat{\theta} \right)}\\
		\alpha &=\left\{\begin{array}{lc}
			1, & \left(h / w \right)>r \\
			2, &  { otherwise }
		\end{array}\right.
	\end{aligned}
\end{equation}
where $h$ and $w$ are the values of the long side  and the short side of the ground truth respectively. 
$r$ is the threshold of the aspect ratio, which is set as $1.5$ in this paper. $\alpha$ is the aspect ratio weighting. 
$\mathcal{L}_{smoothl1}$ is the smooth $\mathcal{L}_1$ loss  \cite{2015Fast9}.
$\theta$ and $\hat{\theta}$ are the rotation angles of ground truth and predition respectively.
If the object has a certain aspect ratio, the period of $\left|\sin\left(\alpha\left(\theta-\hat{\theta} \right)\right)\right|$ is set to $180^{\circ}$, $\alpha=1$; 
Otherwise, if the object is square-like, the period is refined as $90^{\circ}$, $\alpha=2$. 
Therefore, the model is  capable of solving periodicity of angle (POA) problems and adjusting the strategy of trainings more flexibly for bounding boxes with different aspect ratios.

\vspace{-0.6cm}

The loss function of target detection is defined as follows:
\begin{equation}
	\begin{aligned}
		\mathcal{L}_{\mathrm{object}}=& \frac{1}{N} \sum_{n=1}^{N} \sum_{j \in\{x, y, h, w\}} \mathcal{L}_{\mathrm{reg}}\left(t_{n j}, \hat{t}_{n j} \right) \\
		&+\frac{1}{N} \sum_{n=1}^{N} \mathcal{L}_{\theta}\left(\theta_{n},\hat{\theta}_{n} \right)+\frac{1}{N} \sum_{n=1}^{N} \mathcal{L}_{\mathrm{cls}}\left(p_{n}, \hat{p}_{n}\right)
	\end{aligned}
\end{equation}
where $N$ denotes the number of the predicted anchors. $t_{nj}$ is the targets vector of ground truth, and $\hat{t}_{n j}$ denotes the predicted offset vectors.
$p_{n}$ and $\hat{p}_{n}$ is the gournd truth and the probability distribution of various classes respectively. 
The regression loss of detectors $\mathcal{L}_{\mathrm{reg}}$  is the smooth $\mathcal{L}_1$ loss, while the classification loss of detectors $\mathcal{L}_{\mathrm{cls}}$  is the focal loss\cite{2017Focal15}.

\subsection{Denoised  Feature Fusion Module}
The noises in SAR images can disrupt the features extracted by traditional blocks from the input images, thereby making object detection less accurate. 
The strong speckle noise existed in SAR images affects the feature learning by shallow layers, and also hinders the semantic feature learning by higher layers.
The detection performance of SAR ship detection can be improved by the denoised feature maps.
However, the feature information of the input image is significantly reduced by denoising, especially  for small scale targets.
Directly applying the denoised feature map to the following detection module does not perform well on small scale ships, as well as encountering an incorrect detection and missed detection in SAR target detection.  
Therefore, the denoised feature fusion module (DFF) is proposed to obtain the features, which can better represent the target information.  
This module can produce noise-free images directly, but its purpose is not to produce the noise-free image as the input for the segmentation and detection modules, but rather to fusing the original feature map with the denoised feature map.

In general, the DFF module consists of down-sampling convolution, up-sampling convolution and dual-feature fusion attention mechanism. The down-sampling process is composed of six convolutional layers and two max-pooling layers that each reduces the resolution by a factor of two. Then, in the up-sampling stage, the feature map is up-sampled to the size of the original image by the transpose convolution. When the size of the feature map becomes the same as $F_2$, these feature maps are delivered to the DFA mechanism for fusion.

\par
\begin{figure}[H]
	\centering
	\includegraphics[width=1.05\linewidth]{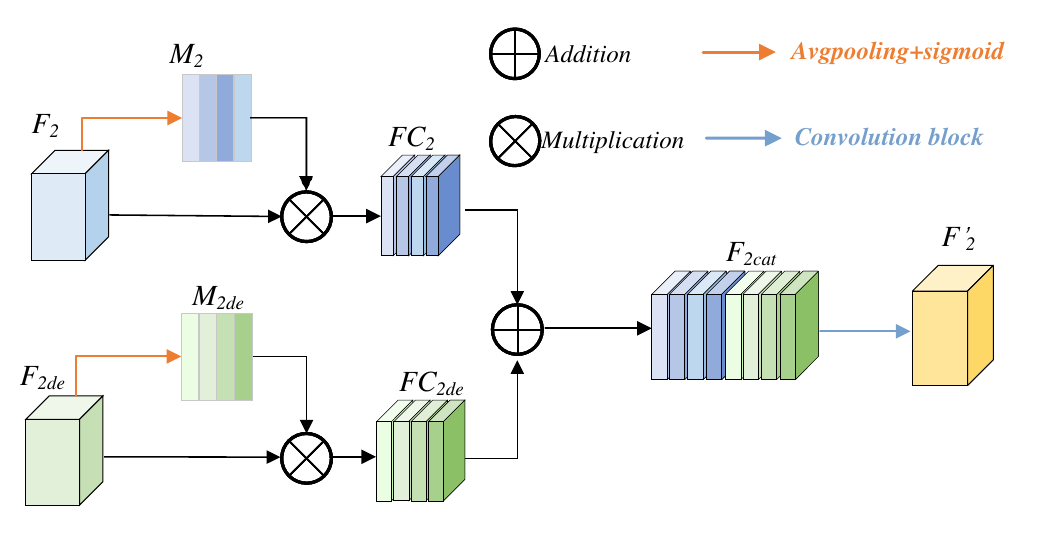}
	\vspace{-0.5cm}
	\caption{Illustration for dual-feature fusion attention  mechanism}
	\label{fig:attention}
\end{figure}

 The dual-feature fusion attention mechanism is illustrated in Fig. \ref{fig:attention}. Take the feature layer $F\in $ $\mathbb{R}^{H \times  W \times C}$ as an example. 
In feature layer $F$, different channels represent different features, and each channel contributes differently to the final detection result. Firstly, two identical average pools are used to compress the spatial dimension of the two feature layers in parallel, obtaining two channel-attention vectors, respectively. Then, these vectors are sent  to a sigmoid function to generate the weight map of channels $M_2$ and $M_{2de}$. Thirdly, the refined feature maps $FC_2$ and $FC_{2de}$ are generated by multiplying the input $F_{2}$, $F_{2de} $, and channel-weight maps $M_{2}$, $M_{2de}$  element by element respectively. Finally, the two feature maps are concatenated and the feature map is obtained with more prominent target information $F_2^{\prime}$ $\in $ $\mathbb{R}^{H \times  W \times C}$ by 1*1 convolution operation. The process can be described as: 
\begin{equation}
F_{2}^{\prime}={f}^{1*1}\otimes\left(M_{2} \otimes F_{2} 
{\oplus} M_{2de} \otimes F_{2de}\right)
\end{equation}
where  ${f}^{1*1}$ is the convolutional layer with 1*1 filters. $\sigma$ denotes the sigmoid function, $\otimes$ is element-wise multiplication,  and $\oplus$  denotes the concatenation of two feature layers. 

\vspace{-0.1cm}
\par The fused features are restored to the same size of the original image by deconvolution. The denoised image is obtained after activating by the sigmoid function. To train clarity enhancement, the denoised feature fusion task adopted the  mean square error (MSE) loss is formulated  as follows:
\begin{equation}
	\mathcal{L}_{\mathrm{denoise}}=\frac{1}{W} \frac{1}{H} \sum_{i=1}^{W} \sum_{j=1}^{H} \left|\left|\mathbf{Y}_{i j}-\hat{\mathbf{Y}}_{i j}\right|\right|^{2}
\end{equation}
where $\hat{\mathbf{Y}}_{ij}$ is the estimated denoised image, and $\mathbf{Y}_{ij}$ is the corresponding ground truth.


\begin{figure*}[hb]
	\centering
	\includegraphics[width=1.02\linewidth]{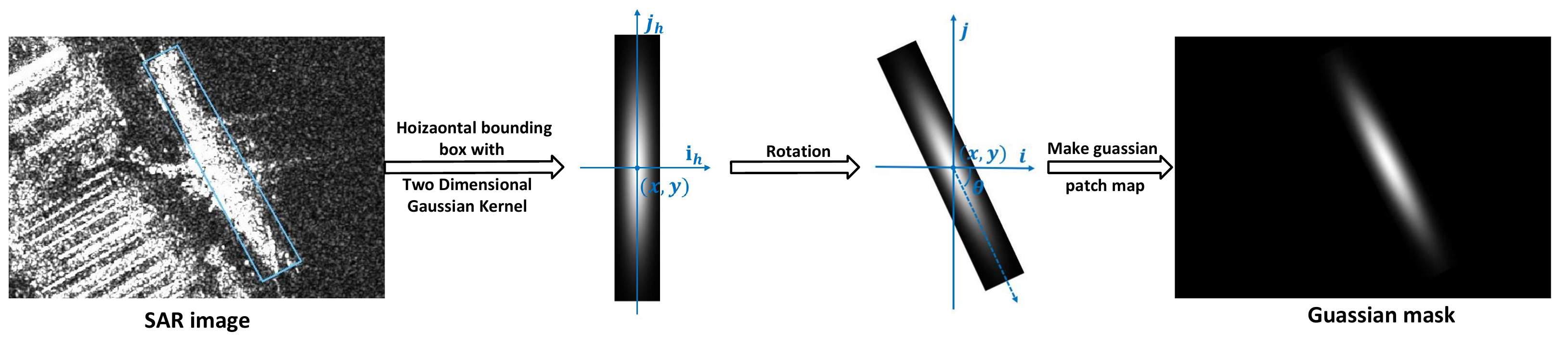}
	\caption{Detailed process of the rotation Gaussian-mask ship labeling.}
	\label{fig:guass}
\end{figure*}

\subsection{Target Segmentation Module}
Since ships have a large aspect ratio, representing ship objects with an oriented bounding box (OBB) often includes a great deal of background information. 
Accordingly, the pixels of the background are distributed at the boundary of the OBB, while the pixels of foreground are concentrated on the center of the OBB, which has a significant impact on ship identification and network convergence.

\par To assist the detection network in extracting the intersection regions from the cluttered background of SAR images, a  target segmentation module with rotated Gaussian-mask is presented for ship modeling based on the geometric characteristics. 
The ship modeling is conducted by assigning the centers of the OBB as the ship centers with the highest confidence, while other pixels are filled according to the gaussian distribution.  
The covariance of confidence distribution is dependent on the aspect ratio of the ships.
As illustrated in Fig. \ref{fig:guass}, the horizontal bounding boxes of the ships are extracted firstly, and meanwhile the long sides and short sides of the ships are respectively defined as the height and width for each bounding box. 
Then, according to the Gaussian distribution rule, the confidence of each pixel in the rotated Gaussian mask can be calculated as follows:

\vspace{-0.4cm}
\begin{equation}
\begin{aligned}
& g(i, j) = \operatorname{exp}\left(-\left(\lambda_{w} \frac{(i_h-x)^{2}}{2 w^{2}}+\lambda_{h} \frac{(j_h-y)^{2}}{2 h^{2}}\right)\right) \\
& \left[\begin{array}{l}
i - x \\
j - y
\end{array}\right]=\left[\begin{array}{cc}
\cos \theta & -\sin \theta \\
\sin \theta & \cos \theta
\end{array}\right] *\left[\begin{array}{l}
i_{h} - x \\
j_{h} - y
\end{array}\right]
\end{aligned}
\end{equation}
where ${(x,y)}$  denotes the coordinate of ship center. 
$ {h}$ and $ {w}$ respectively represent the height and width of the ship objects.
$ {\theta}$ is rotation angle.  
$\lambda_{w}$ and $\lambda_{h}$ are the covariance control factors. 
Depending on the rotation angle $ {\theta}$ of the ship, the coordinate transformation is applied to the coordinates of the horizontal Gaussian mask ${(i_{h},j_{h})}$  and the rotated Gaussian mask ${(i,j)}$.

\par The semantic features are provided by incorporating lower-level feature maps with location information and higher-level feature maps with context information as the outputs of the DFF module.
The dilated convolution offers an efficient mechanism for controlling the field of view, which finds the best balance between accurate localization and context assimilation. 
Therefore, we firstly use two 3*3 convolution layers with different dilated rates to obtain the paired feature maps $<F_{41}, F_{42}>,< F_{51}, F_{52}>$. Then, the small feature map $F_{51}$ and $F_{41}$ are deconvoluted and fused with the large-scale feature map $F_{42}$ and $F_{52}$ by 1*1 convolution to obtain the feature map $F_{3cat}$ and $F_{4cat}$. The same deconvolution and fusion operation is performed on pairs of $<F_{3cat}, F_{4cat}>$ to obtain $F_{se}$. Finally,  the final segmented target is obtained through a series of transpose convolution and sigmoid functions. 

\par As a result of the semantic feature-learning task, the learned features are integrated into the perception of object semantics, thus improving the performance of recognition. The target segmentation loss is formulated as follows:
\begin{equation}
\begin{aligned}
	\mathcal{L}_{\mathrm{segment}}=\frac{1}{W} \frac{1}{H} \sum_{i=1}^{W} \sum_{j=1}^{H} g_{i j}\left(1-\hat{p}_{i j}\right)^{\gamma} \log \left(\hat{p}_{i j}\right)\\
	\hat{p}_{i j}=\left\{
	\begin{array}{rcl}
	p_{i j},     &    & { y_{i j}=1}\\     
	1-p_{i j},   &    & {otherwise}\\
	\end{array} \right.
\end{aligned}
\end{equation}
where $p_{i j} \in\{0,1\}$ is the estimated probability of different classes for the location $(i,j)$,  $y_{i j} \in\{0,1\}$.  
$ y_{i j}=0$ represents the background, and $ y_{i j}=1$ represents the ship. 
$g_{i j}$ is the weight map obtained by the rotated Guassion mask to reduce the background pixel contributions.  $\gamma$ is the hyperparameter of the focal loss.

\subsection{Weighted and Rotated  Boxes Fusion}
\par
\begin{figure}[H]
	\centering
	\includegraphics[width=0.95\linewidth]{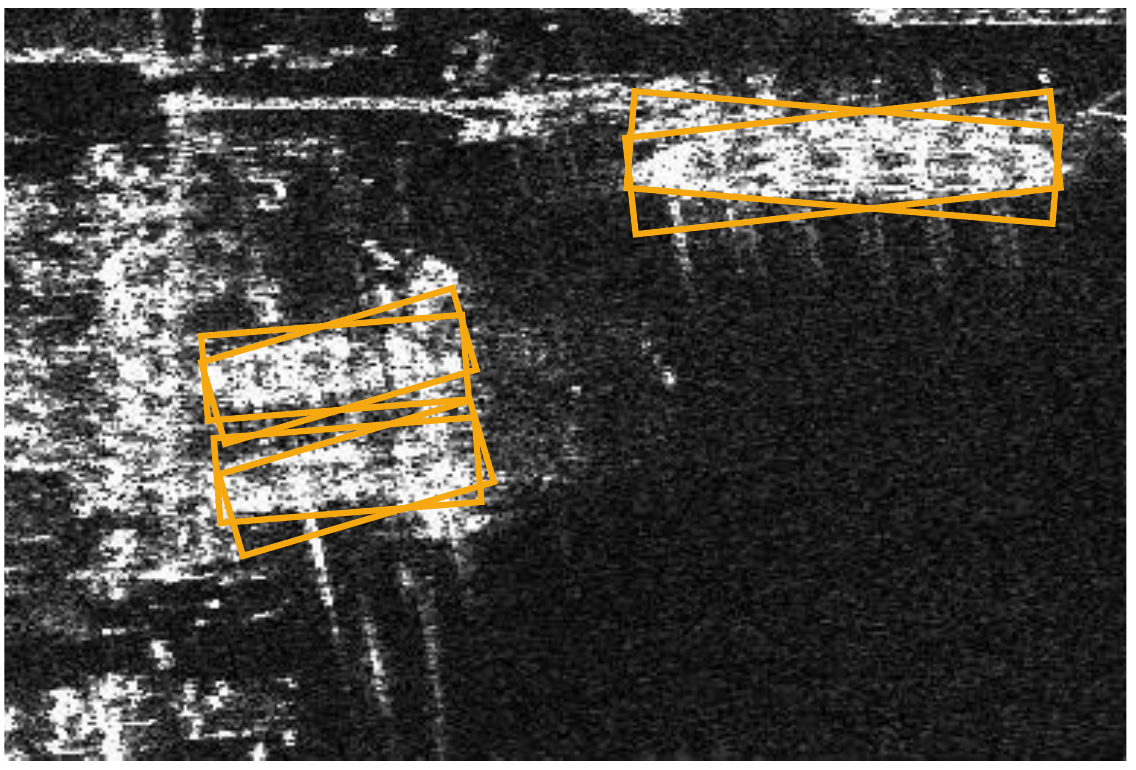}
	\caption{Illustration for anchors  overlapped with each other.}
	\label{fig:WBRF-pic}
\end{figure}

 As shown in Fig. \ref{fig:WBRF-pic}, compared with the original horizontal anchors, the anchors with multiple directions at the same anchor point duplicate by many times, and meanwhile these anchors are  in an overlapped arrangment. 
As well as  anchors beyond the boundary, anchors with low scores as well as high overlap rates also need to be eliminated. 
Non-maximum suppression (NMS) is one of the most commonly used fusion strategies, but it places too much emphasis on classification confidence without considering the accuracy of localization. 
Even though the improved soft-NMS \cite{2017Focal15} can alleviate the existing problem to some extent, many false alarms still occur due to the retention of redundant boxes. 

\par In this paper, expecting to improve the generalization capability of SAR ship detectors, WRBF is adopted to combine the predictions of object detection and target segmentation  to simultaneously take the confidence of classification  as well as the  accuracy of localization into account.
The target segmentation task is not able to directly form the final object OBBs for the following predictions. 
To identify the oriented rectangle with the minimum region that encloses the mask region, the topological structural analysis algorithm \cite{Satoshi1985Topological} is applied. 
\begin{figure}[H]
	\centering
	\includegraphics[width=1\linewidth]{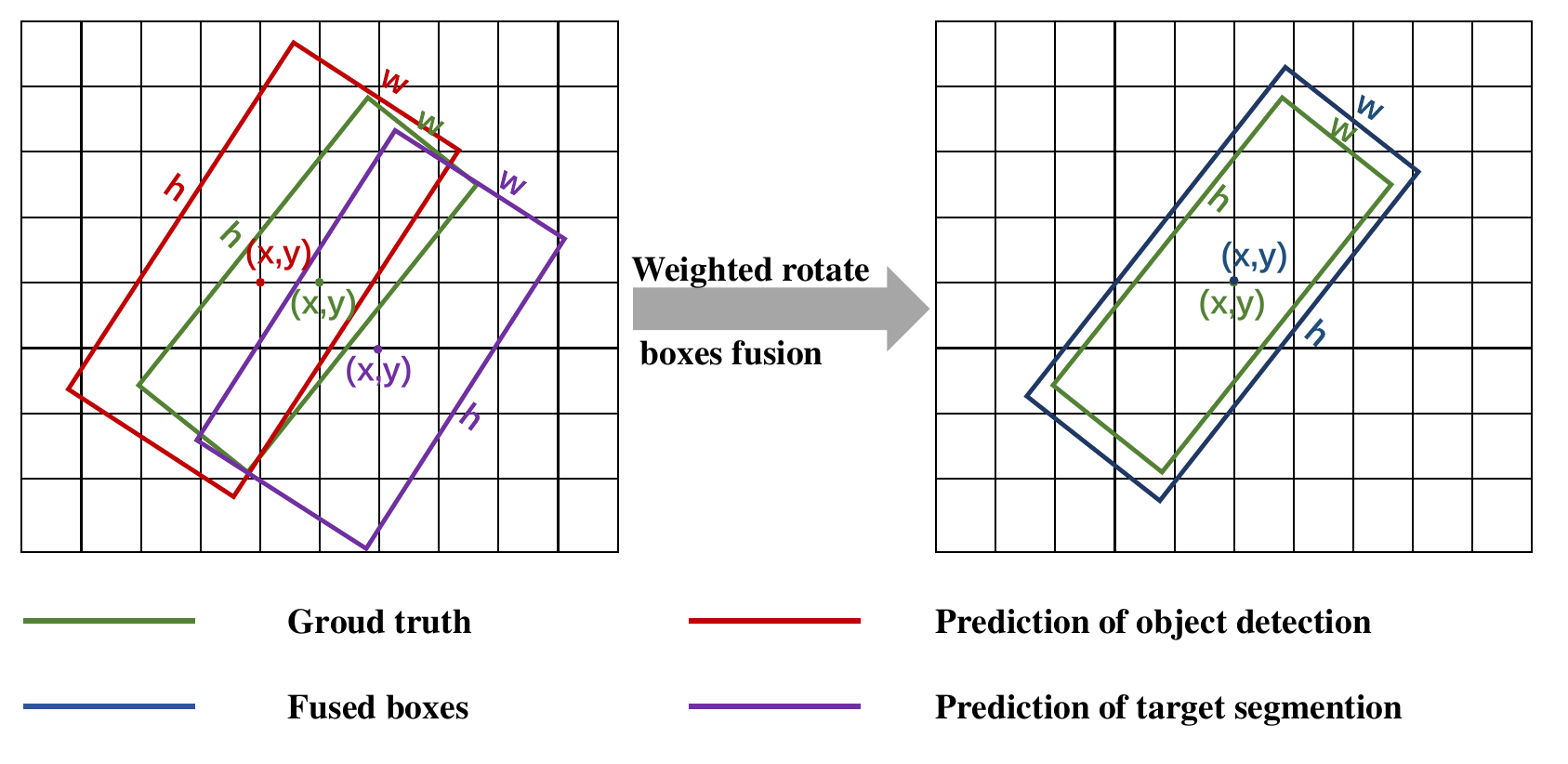}
	\caption{Illustration of the WRBF strategy.}
	\label{fig:fusion boxes}
\end{figure}
According to the WRBF strategy, the confidences  of the predicted boxes are used in order to update the predictions of localizations as shown in Fig. \ref{fig:fusion boxes}. 
To be more specific, the predicted boxes with a higher classification confidence will contribute more proportionately to the final averaged boxes. 
The WRBF strategy is illustrated as Algorithm 1.

\begin{algorithm}[]
	\caption{ Weighted and Rotated Boxes Fusion Strategy.}\label{alg:alg1}
	\begin{algorithmic}[1] 
		\REQUIRE $\mathbf{B_{ob}}$ and $\mathbf{B_{se}}$ are the predicted boxes of object detection task and target segmentation task, respectively. $thr$ is the  threshold of IoU.
		\ENSURE $\mathbf{L_{F}}$ is the fusion boxes.
		\STATE { $\mathbf{L}$ $\leftarrow$ $\mathbf{B_{ob}}$ $\mathbf{B_{se}}$}
		\STATE {Initial: $\mathbf{L_{E}}$: a set of boxes for box clusters. $\mathbf{L_{F}}$: a box on the location.}
		\WHILE {$\mathbf{L}$ is not empty}
		\STATE {
			IoU is calculated between a box in $\mathbf{L_{F}}$ and the  $box_{current}$ in $\mathbf{L}$
		\IF {IoU < $thr$} 
		\STATE {$\mathbf{L^{end}_{E}}$ $\leftarrow$ $box_{current}$,  $\mathbf{L^{end}_{F}}$ $\leftarrow$  $box_{current}$}
		\ELSE
		\STATE {$\mathbf{L^{cur}_{E}}$ $\leftarrow$ $box_{current}$,  $\mathbf{L^{cur}_{F}}$ is updated by $\mathbf{L^{cur}_{E}}$

		\begin{equation}
			\begin{gathered}
				C=\frac{\sum_{n=1}^{T} C_{n}}{T} 
			\end{gathered}
		\end{equation}
		
		\begin{equation}
			\label{eq10}
			\left\{
			\begin{aligned}
				\hat{x}_{n} , \hat{y}_{n} & = \frac{\sum_{n=1}^{T} C_{n} \times \operatorname{\hat{x}_{n}}, \hat{y} _{n}}{\sum_{n=1}^{T} C_{n}}\\
				\hat{h}_{n} ,\hat{w}_{n} & =  \frac{\sum_{n=1}^{T} C_{n} \times \operatorname{\hat{h}_{n}}, \hat{w}_{n}}{\sum_{n=1}^{T} C_{n}}\\
				\cos\hat{\theta}_{n} ,\sin\hat{\theta}_{n} & = \frac{\sum_{n=1}^{T} C_{n} \times \operatorname{\cos\hat{\theta}_{n}}, \sin\hat{\theta}_{n}}{\sum_{n=1}^{T} C_{n}}\\
			\end{aligned}
			\right.
		\end{equation}
	}
		\ENDIF
	}
		\ENDWHILE

	\end{algorithmic} 
	\label{alg1}
\end{algorithm}

\par 
The joint training loss  is given as inluding the losses for denoise, segmentation, detection and classification.
\begin{equation}
	\mathcal{L}_{\mathrm{total}} = \lambda_{1}*\mathcal{L}_{\mathrm{denoise}}+\lambda_{2}*\mathcal{L}_{\mathrm{segment}}+\lambda_{3}*\mathcal{L}_{\mathrm{object}}
\end{equation}
where $\lambda_{1}, \lambda_{2},\lambda_{3}$ are  the hyper-parameters of object detection subtask, denoising subtask and target segemention subtask respectively.

\section{EXPERIMENTS}
\subsection{Experimental Data and Platform} 

\par 
In this experiment, the performance of our proposed method is evaluated and analyzed using two representative SAR ship datasets, i.e., SAR ship detection dataset+ (SSDD+) and High-Resolution SAR Images Dataset (HRSID). 
SSDD+, which is a multisensor, multiresolution, and multisize dataset provided by Li et al. \cite{2017ShipSSDD}.
There are 1160 images in the SSDD dataset and 2456 ships ranging in scale from 7  ${\times}$ 7 to 211  ${\times}$ 298.
These images are available in the following polarization modes: HH, HV, VV, and VH, with the different resolution from 1 to 15 meters. 
The HRSID provided by Wei et al. \cite{2020HRSID} is established by  images from 36 TerraSAR-X images, 99 Sentinel-1B images, and 1 TanDEM-X image for ship instance segmentation and ship detection. 
As a result of the cropping process, these large-scale images have been cropped out into 800  ${\times}$ 800 pixels. 
Consequently, it includes  16951 ships distributed in 5604  SAR images in total. 
These cropped images are divided into  the training set, the validating set, and the testing set  with the proportion of 7:2:1.
As for the benchmark, the images in two datasets are resized to 320 ${\times}$ 320 pixels, which serves as the inputs of detectors.
In the denoising task, since that there are usually no corresponding noise-free images for real SAR images,  we adopt the training idea in \cite{tan2021cnn} and \cite{2022Transformer} to treat the two corresponding  SAR images with the same scene and different polarization methods as the label pairs for denoising network training. 
137327 pairs of corresponding  HH and HV images are selected as the training set for DDF module.

\subsection{Evaluation Metrics} 
Precision, recall and F1 scores are employed to evaluate the performance of SAR ship detectors. The definition of these evaluation metrics is given as follows:
\begin{equation}
	\begin{gathered}
		\text { Precision }=\frac{\mathrm{N}_{\mathrm{TP}}}{\mathrm{N}_{\mathrm{TP}}+\mathrm{N}_{\mathrm{FP}}} 
	\end{gathered}
\end{equation}
\begin{equation}
	\begin{gathered}
		\text { Recall }=\frac{\mathrm{N}_{\mathrm{TP}}}{\mathrm{N}_{\mathrm{TP}}+\mathrm{N}_{\mathrm{FN}}} \\
	\end{gathered}
\end{equation}
where $\mathrm{N_{TP}}$ (true positives), $\mathrm{N_{FP}}$ (false positive), and $\mathrm{N_{FN}}$ (false negative) refer to the numbers of correctly detected ships, false alarms, and missing ships respectively. 
A predicted bounding box is considered as a true positive if its IoU with the ground truth is higher than a given IoU threshold, i.e., 0.5 in this paper. Otherwise, it is regarded as a false positive. Moreover, the predicted bounding boxes with the highest confidence score are seen as the true positive, if the IoU of several ones with the ground truth are all higher than the threshold. 

F1-score is a comprehensive evaluation metric for the quantitative performance of different models by simultaneously considering the precision rate and recall rate. 
\begin{equation}
	\begin{gathered}
		\text {F1-score} =\frac{2 \times \text { Precision } \times \text { Recall }}{\text { Precision }+\text { Recall }}
	\end{gathered}
\end{equation}

Additionally, the average precision (AP) metric is the most frequently used metric to evaluate the performance of a detector. This paper only calculates the AP of the ships to assess the ability of our method. AP is defined as the area under the precision–recall curve as follows:

\begin{equation}
	\mathrm{AP}=\int_{0}^{1} P(R) d R
\end{equation}
If $\mathrm{AP}$ is calculated at the IoU threshold of 0.5, it can be defined as $\mathrm{AP_{50}}$. Similarly, $\mathrm{AP_{75}}$ denotes the $\mathrm{AP}$ calculated at the IoU threshold of 0.75, which needs higher localization accuracy to further evaluate the detection ability. $\mathrm{AP_{75}}$ is an evaluation indicator for the MS COCO dataset \cite{2014arXiv1405}.

\begin{table*}[th]
	\centering
	\setlength{\tabcolsep}{6mm}
	\caption{ABLATION STUDIES OF MLDet ON \textbf{SSDD+}}
	\label{fig: table SSDD+}
	\begin{tabular}{ccccccccc}
		\toprule[0.3mm]
		DFF Module       &  TS Module      &ARW loss    & Precision & Recall & F1-score & $Ap_{50}$ & $AP_{75}$   \\
		\midrule[0.1mm]
		\XSolidBrush	&\XSolidBrush	&\XSolidBrush	&	0.839 	&	0.825 	&	0.832 	&	0.847 	&	0.475 	\\
		\CheckmarkBold	&\XSolidBrush	&\XSolidBrush	&	0.853 	&	0.866 	&	0.859 	&	0.893 	&	0.502 	\\
		\XSolidBrush	&\CheckmarkBold	&\XSolidBrush	&	0.916 	&	0.895 	&	0.905 	&	0.917 	&	0.579 	\\
		\XSolidBrush	&\XSolidBrush	&\CheckmarkBold	&	0.855 	&	0.830 	&	0.842 	&	0.864 	&	0.492 	\\
		\CheckmarkBold	&\CheckmarkBold	&\XSolidBrush	&	0.933 	&	0.945 	&	0.939 	&	0.947 	&	0.567 	\\
		\CheckmarkBold	&\XSolidBrush	&\CheckmarkBold	&	0.872 	&	0.888 	&	0.880 	&	0.918 	&	0.540 	\\
		\XSolidBrush	&\CheckmarkBold	&\CheckmarkBold	&	0.937 	&	0.940 	&	0.939 	&	0.939 	&	0.565 	\\
		\CheckmarkBold	&\CheckmarkBold	&\CheckmarkBold	&	\textbf{0.941}	&	\textbf{0.946}	&	\textbf{0.943}	&	\textbf{0.953}		&	\textbf{0.601}	\\
		\bottomrule[0.3mm]
	\end{tabular}
\end{table*}

\begin{table*}[th]
	\centering
	\setlength{\tabcolsep}{6mm}
	\caption{ABLATION STUDIES OF MLDet ON \textbf{HRSID}}
	\label{fig: table HRSID}
	\begin{tabular}{ccccccccc}
		\toprule[0.3mm]
		DFF Module       &  TS Module      &ARW loss    & Precision & Recall & F1-score & $Ap_{50}$ & $AP_{75}$   \\
		\midrule[0.1mm]
		\XSolidBrush	&\XSolidBrush	&\XSolidBrush	&	0.817 	&	0.810 	&	0.813 	&	0.814 	&	0.452 	\\
		\CheckmarkBold	&\XSolidBrush	&\XSolidBrush	&	0.856 	&	0.837 	&	0.847 	&	0.855 	&	0.492 	\\
		\XSolidBrush	&\CheckmarkBold	&\XSolidBrush	&	0.909 	&	0.876 	&	0.892 	&	0.883 	&	0.521 	\\
		\XSolidBrush	&\XSolidBrush	&\CheckmarkBold	&	0.838 	&	0.820 	&	0.829 	&	0.824 	&	0.438 	\\
		\CheckmarkBold	&\CheckmarkBold	&\XSolidBrush	&	0.910 	&	0.902 	&	0.906 	&	0.917 	&	0.556 	\\
		\CheckmarkBold	&\XSolidBrush	&\CheckmarkBold	&	0.884 	&	0.862 	&	0.873 	&	0.868 	&	0.515 	\\
		\XSolidBrush	&\CheckmarkBold	&\CheckmarkBold	&	0.916 	&	0.903 	&	0.909 	&	0.896 	&	0.524 	\\
		\CheckmarkBold	&\CheckmarkBold	&\CheckmarkBold	&	\textbf{0.931} &	\textbf{0.915}	&	\textbf{0.923}	&	\textbf{0.928} 	&	\textbf{0.534}	\\
		\bottomrule[0.3mm]
	\end{tabular}
\end{table*}

\subsection{Experiment Settings} 
The experiments are conducted using the Pytorch framework with Ubuntu 16.04 and Python 3.5.
A NVIDIA RTX2080Ti GPU and 128GB memory are included in the experimental hardware platform.
Throughout this paper, the ablation experiments and parameter analysis  have been conducted using the SAR datasets mentioned above. 
The network architecture CSPDArknet in the single-stage yolo series is selected  as the backbone to effectively extract features. 
Pre-processing the dataset is important to save memory and speed up training while improving training accuracy. For both datasets, some pre-processing skills are used in experiments on different models, including scaling the images to the same size, normalising the pixel values, data enhancement.

Training a detection model can be done without the need for large amounts of data and pre-trained models\cite{2017arXiv170801241S}, especially for the modified structures of pre-trained models and specific image datasets.
Therefore,  our model is trained  from scratch, meaning that no pre-trained weights are loaded. 
The training process consists of two stages. 
In the first training stage,  the proposed network is initialized using Gaussian random variables, and the DDF module is independently trained by the dataset SSDD+. 
In this stage, the initial learning rate is set as 0.001, and the momentum is set as 0.9. The weight decay and  the training epoch are respectively set as 0.0005 and 50. 
The training speed and accuracy of the model are very easily affected by the learning rate setting. 
Therefore, the learning rate scheduling strategy with cosine annealing is employed as the learning rate adjustment method \cite{2017arXiv170801241S}.
The learning rate is scaled according to the cosine function value from 0 to PI, making the training more stable.
$\lambda_{1}$ is set to 1, while $\lambda_{2}$ and $\lambda_{3}$ are set to 0 for the loss function to train the DDF module individually.
In the second training stage (the latter 50 epochs during the unfrozen  training in our experiment), the  parameters in the DDF module is freezed, and  a lower learning rate is initialized as 0.0001.
The other settings in the second traning stage are as the same as the first stage. 
$\lambda_{1}$, $\lambda_{2}$ and $\lambda_{3}$ in the loss  function are set as 0, 1, and 1 respectively.  
The parameters of the OB module and the TS module are learned simutanously in the second training stage.
Meanwhile, the commonly used training techniques such as batch normalisation, residual concatenation, dropout are also adopted in experiments on different model training, which help the models to fit the dataset better.

\subsection{Evaluation of Our Proposed Method}
To demonstrate the effectiveness for each modification, a range of ablation experiments will be conducted to quantify the impact of each improvement on SSDD+ and HRSID. A detailed analysis of each improvement for the network is then conducted following the discussion. 

\subsubsection{Ablation Experiments Results}

\par
\begin{figure}[H]   
	\centering         
	\includegraphics[width=1\linewidth]{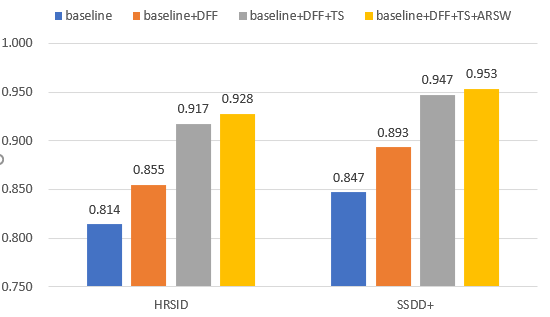}
	\caption{Comparison of experiments containing multiple improvements on HRSID (left) and SSDD+ (right).
		The blue, orange, grey and yellow bar charts represent the experimental results ($Ap_{50}$) of baseline, baseline with DDF module, baseline with DDF module and TS module, and MLDet (baseline with DDF module, TS module and ARW loss), respectively.}
	\label{fig:Model performance}
\end{figure}

 Since a number of improvements have been made to the model design and training process, it is necessary to examine the actual effects of each improvement in the context of each other.
Different combinations of these improvements were tested, and the results of these experiments are presented in Tables \ref{fig: table SSDD+} and Table \ref{fig: table HRSID} respectively. 
Based on the experimental results, the following aspects can be summarized.
Firstly, it can be observed from  Table \ref{fig: table SSDD+} and Table \ref{fig: table HRSID} that the resutls of applying each improvement of our method individually   demonstrates the effectiveness  with respect to the benchmark model. 
Secondly, as shown in Fig. \ref{fig:Model performance}, compared with the experiments including only one improvement, experiments including multiple improvements demonstrate a significant cumulative performance gain with a small overlap between them.

\par Compared with the baseline, MLDet achieves 10.6$\%$ higher  $AP_{50}$ on SSDD+ and  11.4$\%$ higher $AP_{50}$ on HRSID respectively. 
In addition, when the accuracy for localization  increases and the IoU threshold is set to 0.75, the $AP_{75}$ gains a larger improvement of 12.5$\%$ on SSDD+ and  10.2$\%$ on HRSID, which  indicates that the bounding boxes predicted by the MLDet are more accurate. The performance of each component will be discussed  in detail as the followings.

\subsubsection{Effect of The Denoised  Feature Fusion Module} 

\par
\begin{figure}[h]
	\centering
	\includegraphics[width=1.02\linewidth]{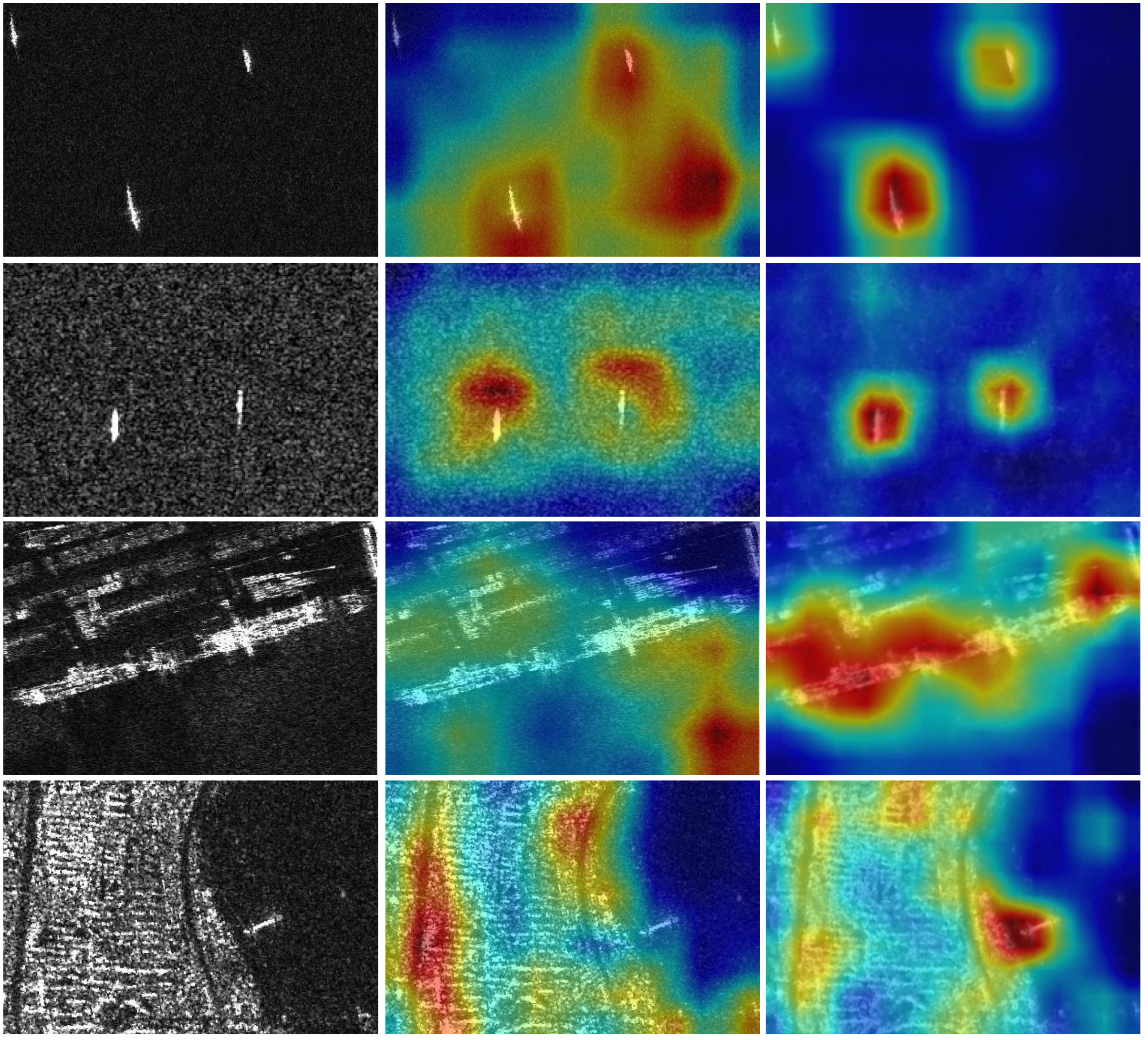}
	\caption{The results of the dual-feature fusion attention mechanism enables our MLDet to focus on both inshore and offshore ships. The left  is the original image, the middle  is the heat map without dual-feature fusion attention mechanism, and the right is the heat map after dual-feature fusion attention mechanism.}
	\label{fig:hotmap}
\end{figure}

 Speckle noise existed in the area of the port is very likely to cause the false alarms than the speckle noise existed in the area of the sea. 
The bright speckles on the port  are easily to be mistakenly identified as ships by many detection models.
Although denoising mechanism can filter the redundant information, the small ship targets including fewer pixels may be filtered out during speckle suppression. 
In the proposed method, the attention mechanism is adopted into the DFF module to simultaneously integrate the denoised features and original features, which benefits the subsequent detection task and segmentation task, especially for the small ship targets.
The dual-feature fusion attention mechanism is based on the binary classification of the ship class and background class, by using CAM (class activation map) \cite{2016Learning}.  The feature map obtained by DFF module does not only eliminate the features of speckle noise, but also preserves the features of small targets.
Besides that, it can be seen from the comparisons in Fig. \ref{fig:hotmap}, the dual-feature fusion attention mechanism greatly activates attention to the ship regions.

\begin{figure*}[hb]
	\centering
	\includegraphics[width=1\linewidth]{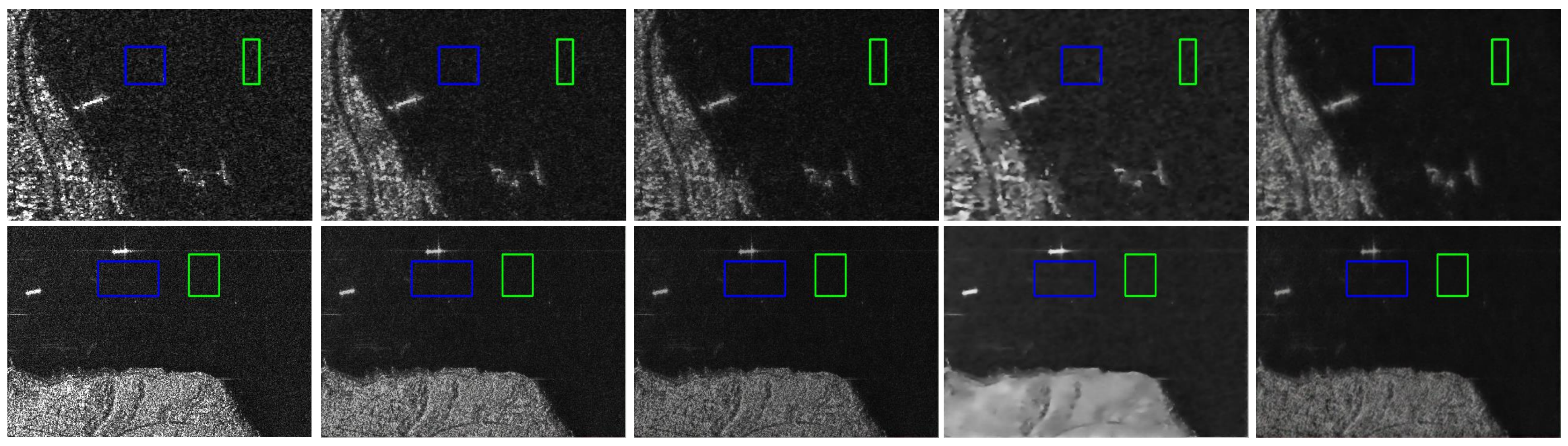}
	\begin{minipage}[t]{1\linewidth}      
		\vspace{0cm} 
		\qquad  Real SAR image  \qquad  \qquad SAR-BM3D \qquad  \qquad Trans-SAR   \qquad  \qquad  \qquad  SAR-ON \qquad \qquad  \qquad \quad MLDet 
	\end{minipage}
	
	\caption{Qualitative results of different despeckling methods on real SAR-1 images. The area represented by the blue rectangle in the first row is marked as region 1, the green rectangle is marked as region 2, the area represented by the blue rectangle  in the second row is marked as region 3, and the green rectangle is marked as region 4.}
	\label{fig:denoise-all}
\end{figure*}

In order to demonstrate the effectiveness to speckle noise, we selected a large number of SAR images from SSDD+ dataset. 
The denoised feature learning task assists MLDet in learning more discriminative targets and features, and  also enables our method to recognize the targets from the severe speckle noise SAR images with a higher degree of confidence. 
As shown in the first and the second row of Table I and II, DFF module achieves 89.3$\%$ on $AP_{50}$ and improves the baseline by approximately five points on SSDD+, and achieves 85.5$\%$ on  $AP_{50}$.  
As a result, the modified network exhibits a higher level of adaptive feature selection, as well as a higher complementarity between features.

\subsubsection{ Effect Of The Target Segmentation Module}
\par Assisted by the rotated Gaussian-Mask, we observe  8$\%$ and 6.9$\%$ increase in $AP_{50}$ as shown in the third row of Table \ref{fig: table SSDD+} and Table \ref{fig: table HRSID} respectively. 
By solving the problem of mislabeling, the discriminative capability of the detector is enhanced, which has a great impact on achieving better detection accuracy.
Furthermore, assited with utilizing the context information, 7$\%$ and 6.6$\%$ improvement on recall is obtained, respectively. 
With adaptive coordinate attention, TS module is able to extract more precise information spatial locations, as well as obtain much more representative feature maps.
Therefore, the network is able to pay more attention on the ship objects amidst the interference of the complex backgrounds. 

Furthermore, the WBRF strategy adjusts both the confidence of the predicted boxes as well as their positions, so as to improve the accuracy of localizations. 
As shown in Table \ref{fig:SSDD+ WFBR} and Table \ref{fig:HRSID WFBR}, 3.6$\%$ and 2.5$\%$ improvements on $Ap_{50}$ are obtained by utilizing the WBRF strategy, respectively. In essence, the TS module is capable of reducing the impact of complex backgrounds on detection performance and enabling accurate target positioning. Additionally, the ablation study also confirms the effectiveness of this procedure.

\begin{table}[h]
	\caption{ABLATION STUDIES OF WBRF STRATEGY  ON \textbf{SSDD+}}
	\centering
	\setlength{\tabcolsep}{2.1mm}
	\begin{tabular}{lcccccc}
		\toprule[0.1mm]
		Methods       & Precision & Recall & F1-score &  $Ap_{50}$  &$Ap_{75}$\\
		\midrule[0.1mm]
		baseline       & 0.839     & 0.825  & 0.832    & 0.847      &0.475 \\
		MLDet-w/o WBRF & 0.884     & 0.873  & 0.878    & 0.881      &0.593 \\
		MLDet-w WBRF   & \textbf{0.916}     & \textbf{0.895}  & \textbf{0.905}    & \textbf{0.917}      &\textbf{0.625} \\
		\bottomrule[0.1mm]
	\end{tabular}
	\label{fig:SSDD+ WFBR}
\end{table}

\begin{table}[h]
	\caption{ABLATION STUDIES OF WBRF STRATEGY ON \textbf{HRSID}}
	\centering
	\setlength{\tabcolsep}{2.1mm}
	\begin{tabular}{lcccccc}
		\toprule[0.1mm]
		Methods        & Precision & Recall & F1-score  & $Ap_{50}$ &$Ap_{75}$ \\
		\midrule[0.1mm]
		baseline       & 0.817 	&	0.810 	&	0.813 	&	0.817   &0.452 \\
		MLDet-w/o WBRF & 0.864	&	0.868	&	0.866 	&	0.858   &0.513 \\
		MLDet-w WBRF   & \textbf{0.909} 	&	\textbf{0.876} 	&	\textbf{0.892} 	&	\textbf{0.883} 	& \textbf{0.548} \\
		\bottomrule[0.1mm]
	\end{tabular}
	\label{fig:HRSID WFBR}
\end{table}

\begin{figure*}[!ht]
	\centering
	\includegraphics[width=1\linewidth]{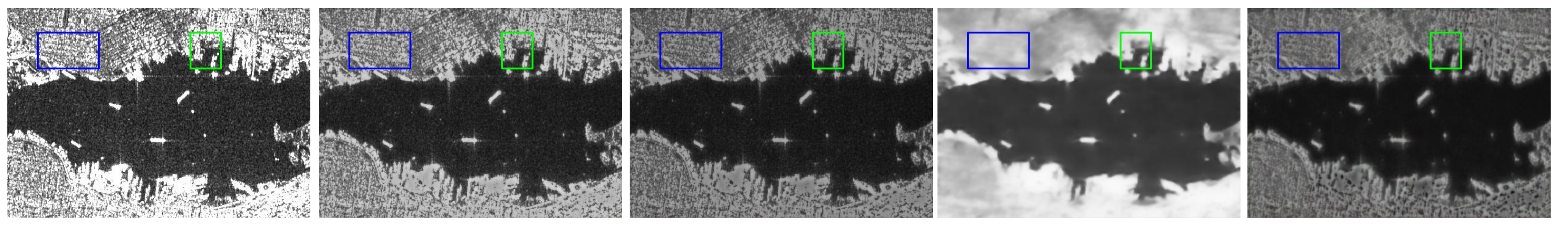}
	\begin{minipage}[t]{1\linewidth}      
		\vspace{0cm} 
		\qquad  Real SAR image  \qquad  \qquad SAR-BM3D \qquad  \qquad Trans-SAR   \qquad  \qquad  \qquad  SAR-ON \qquad \qquad  \qquad \quad MLDet 
	\end{minipage}
	
	\caption{Qualitative results of different despeckling methods on real SAR-2 images. Other methods over-smooths the SAR image, destroying edges and structural details. The area represented by the blue rectangle in the first row is marked as region 5, and the green rectangle is marked as region 6.}
	\label{fig:denoise-worse}
\end{figure*}

\subsubsection{Effect of the ARWS Loss}
\par The periodicity of angular problem  increases the difficulty of rotating object detection. 
A sharp increase in the loss value will be caused in the case that the aspect ratio between the ground truth and the candidate prediction bounding boxes is close to 1, which also brings difficulties for the network trainings. 
Compared with the baseline, the mAP value of the ship targets is obviously improved by the proposed ARWS loss, especially for square-like boxes. The ARWS loss function yields a higher $Ap_{50}$ and $AP_{75}$  than the traditional loss function on SSDD+ and HRSID. 
As a result, our regression method is capable of effectively avoiding a sharp increase in the loss value during the network training, which is advantageous to the detection performance.

\subsection{Comparison With Other Denoised Methods}

We evaluated the performance of the proposed method and recent state-of-the-art methods on real SAR images, including SAR-BM3D\cite{2012A}, Trans-SAR\cite{2022Transformer} and SAR-ON\cite{2022arXiv220515906P}.  
Visual inspection is an important way to qualitatively evaluate the performance of denoising methods in the cases that  noise-free references are unavailable. The denoising results corresponding to the real images are shown in Fig.  \ref{fig:denoise-all} and  Fig. \ref{fig:denoise-worse}. These results demonstrate that our MLDet method can achieve much better performance on suppressing  the speckle noise for  SAR images.

\begin{table}[h]
	\caption{ESTIMATED \textbf{ENL VALUES} ON REAL SAR-1 IMAGES}
	\centering
	\setlength{\tabcolsep}{4mm}
	\begin{tabular}{lcccc}
		\toprule[0.1mm]
		Method         & region1 & region2 & region3 & region4 \\
		\midrule[0.1mm]
		SAR-BM3D        & 74.89  & 85.96   & 10.59   & 11.73    \\
		Trans-SAR       & 81.39  & 85.26   & 184.69  & 212.95   \\
		SAR-ON          & 106.64 &137.14   & 178.32   &  250.10   \\
		MLDet (Ours)    &  \textbf{167.63} & \textbf{197.24}  &  \textbf{262.08} & \textbf{ 263.83}  \\
		\bottomrule[0.1mm]
		\label{fig:ENL}
	\end{tabular}
\end{table}

\begin{table}[h]
	\caption{ESTIMATED \textbf{EPD-ROA VALUES} ON REAL SAR-2 IMAGES}
	\centering
	\setlength{\tabcolsep}{4mm}
	\begin{tabular}{lcccc}
		\hline
		\multicolumn{1}{c}{\multirow{2}{*}{Method}} & \multicolumn{2}{c}{region5}       & \multicolumn{2}{c}{region6}       \\ \cline{2-5} 
		\multicolumn{1}{c}{}                        & HD              & VD              & HD              & VD              \\ \hline
		SAR-BM3D                                    & 0.8778          & 0.6800          & 0.731           & 0.7123          \\
		Trans-SAR                                   & 0.8342          & 0.8923          & 0.9123          & 0.9314          \\
		SAR-ON                                      & 0.6523          & 0.7631          & 0.752           & 0.826           \\
		MLDet (Ours)                                & \textbf{0.9742} & \textbf{0.9134} & \textbf{0.9532} & \textbf{0.9515} \\ \hline
		\label{fig:EPD-ROA}	
	\end{tabular}
\end{table}

In addition, we select several indicators  that do not require noise-free images as reference for
evaluation, namely equivalent number of looks (ENL) \cite{1981Speckle}, edge preservation degree based on the ratio of average (EPD-ROA) in the horizontal direction  (HD) and vertical direction (VD) \cite{2016SAR_EPD}. ENL is used to measure the relative strength of the speckle noise in  images and filter performance. The larger the ENL value, the smoother the homogeneous area, and the better the filtering effect. EPD-ROA is used to measure the detail preservation ability of denoised images. 
When the denoising method preserves much more details, the value EPD-ROA of the denoised image is much closer to 1. 
The ENL values are estimated from the homogeneous regions (shown with blue and green boxes in  Fig.  \ref{fig:denoise-all}). The ENL results are also tabulated in Table \ref{fig:ENL}.  It can be observed from these results that the proposed MLDet outperforms the others compared methods in all four homogeneous regions and signifies the best despeckling performance out of the considered approaches. It also can be seen from Fig. \ref{fig:denoise-worse} and Table \ref{fig:EPD-ROA} that, other methods over-smooth the SAR image, destroying edges and structural details to some extent. Ours method retains more textural details while minimizing distortions in homogeneous regions.
\begin{figure}[!h]
	\centering
	\includegraphics[width=1.02\linewidth]{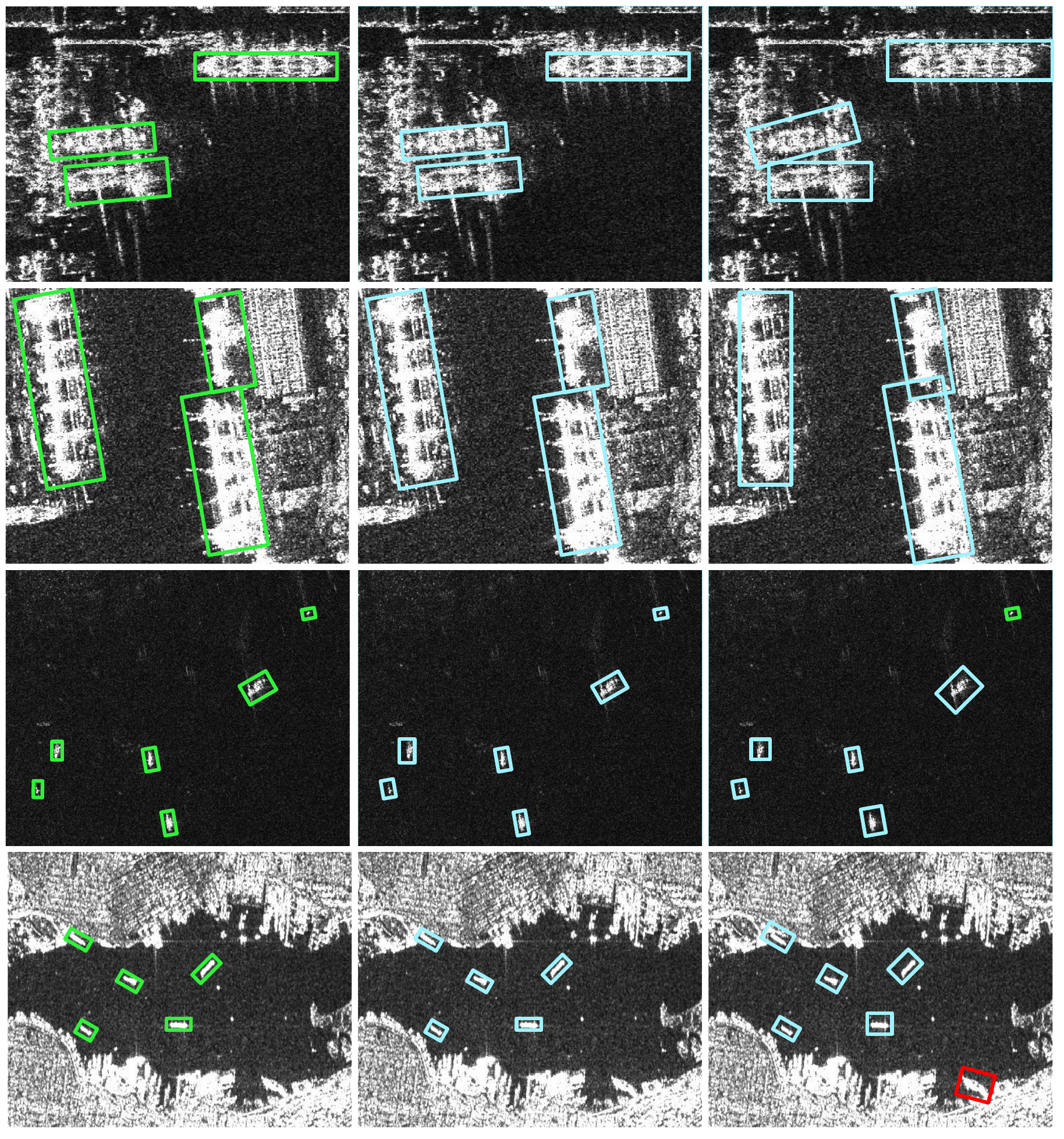}
	\begin{minipage}[t]{1\linewidth}      
		\vspace{0cm} 
		\qquad \qquad(a) \qquad \qquad \qquad (b) \qquad \qquad \quad (c)  
	\end{minipage}
	
	\caption{Comparison of  representative detection results by the proposed  MLDet and the MLDet in sequential form. (a) The ground-truth. (b) MLDet (our method). (c) MLDet in sequential form. Note that the blue boxes are true positive ship targets, the red boxes are false positive ship targets, and green boxes are missed ship targets.}
	\label{fig:part-result}
\end{figure}
\begin{figure*}[h]
	\centering
	\includegraphics[width=1.02\linewidth]{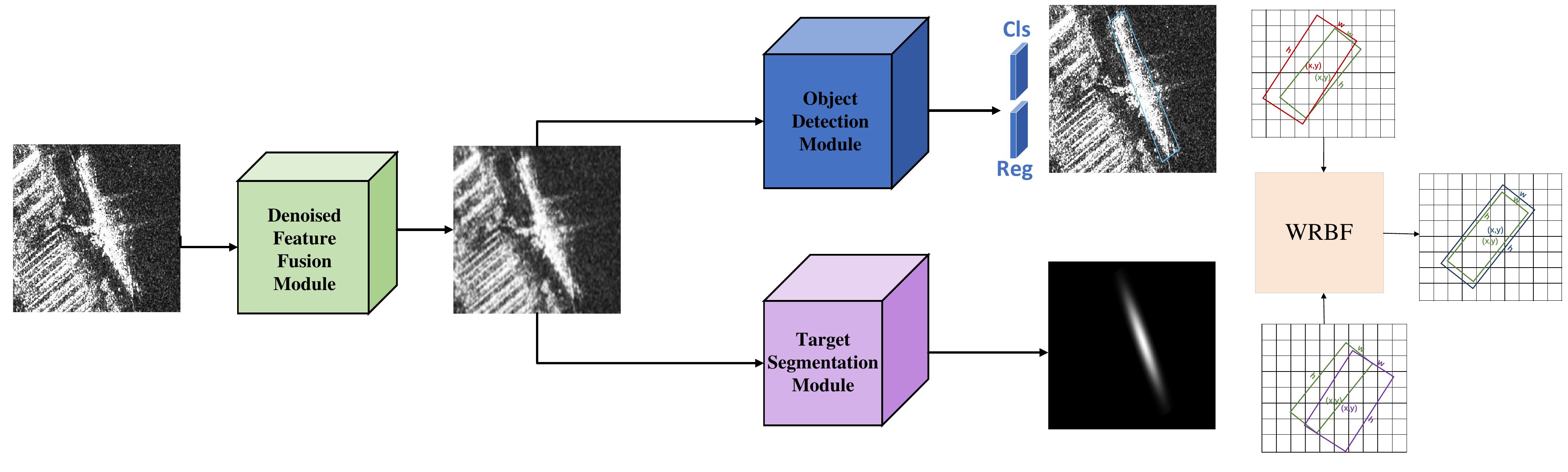}
	\caption{MLDet in sequential form. Firstly, it performs the denoising task, use the L2 loss function to process the Sar image with strong multiplicative noise, and get a relatively clean image, then use the backbone network that does not share the weight, execute detection task and segmentation task respectively,  and finally use the WRBF strategy to fuse the different task results according to different weights to get the final result.}
	\label{fig:part}
\end{figure*}

\begin{figure*}[h]   
	\centering 
	\subfloat[SSDD+]{
		\includegraphics[width=0.4\linewidth]{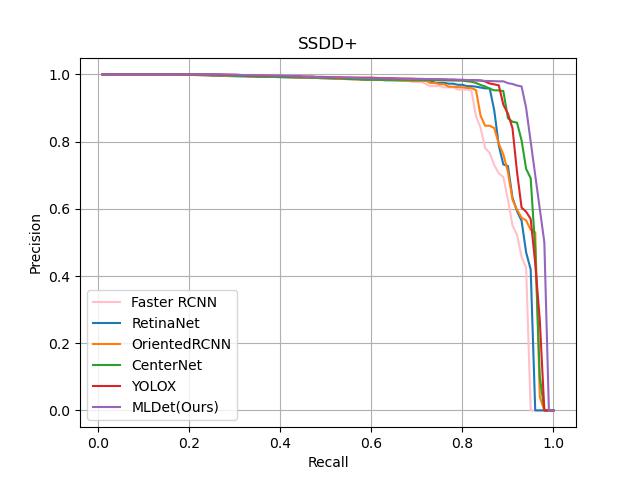}
		\label{fig:SSDD+ PR}
	}
	\subfloat[HRSID]{
		\includegraphics[width=0.4\linewidth]{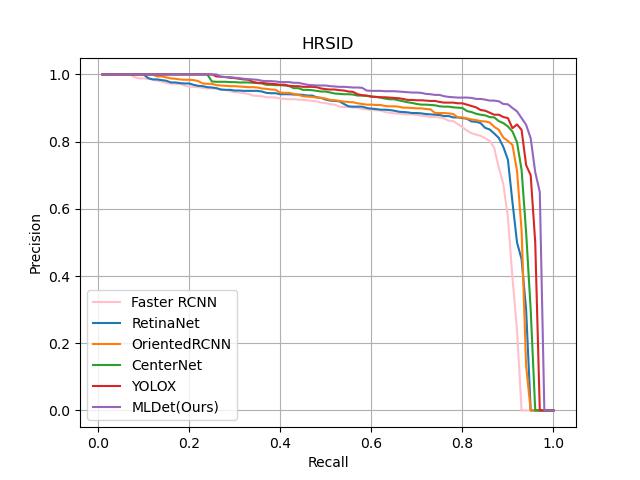}
		\label{fig:HRSID PR}
	}
	\caption{Precision-recall curves of six methods: Faster R-CNN, RetinaNet, Oriented R-CNN, CenterNet, YOLOX and MLDet on (a) SSDD+ and (b) HRSID under the metric of $Ap_{50}$  }
	\label{fig:PR}
\end{figure*}

\begin{figure*}[th]
	\centering
	\includegraphics[width=1\linewidth]{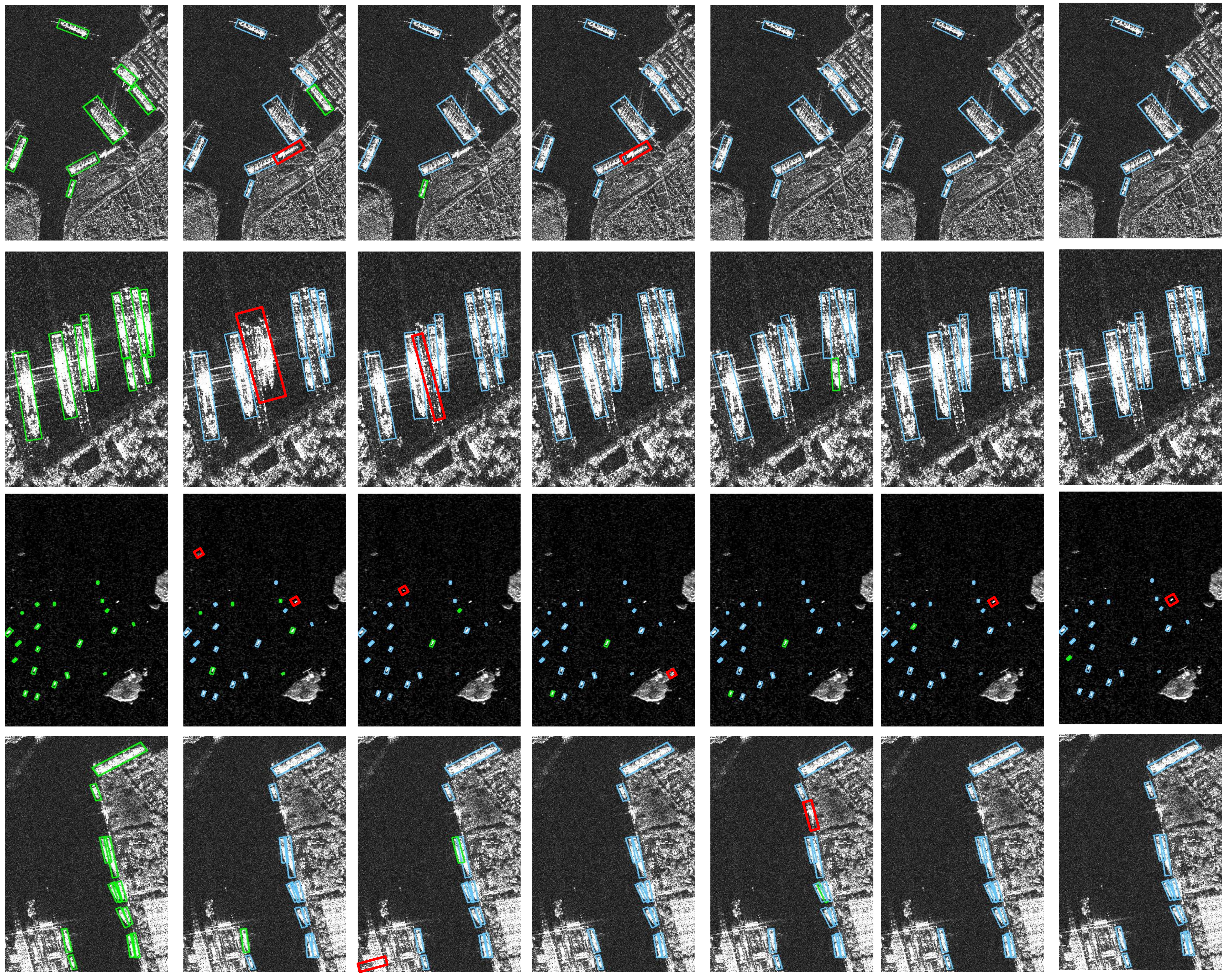}
	\begin{minipage}[t]{1\linewidth}      
		\vspace{0cm} 
		\qquad \qquad(a) \qquad \qquad \qquad (b) \qquad \qquad \quad (c)  \qquad  \qquad \qquad (d) \qquad \qquad \quad\quad (e) \qquad \qquad \quad\quad (f) \qquad \qquad \quad\quad (g)
	\end{minipage}
	
	\caption{Comparison of detection results of different methods on SSDD+. (a) The ground truth. (b) Faster RCNN. (c) RetinaNet. (d) Oriented RCNN. (e) CenterNet. (f) YOLOX. (g) MLDet (our method). The first line shows  the detection results under complex background, the second line shows the detection results under  densely arranged ships, the third line shows the detection results  for small ships, and the fourth line shows the detection results for multiscale  ships. Note that the blue boxes are true positive ship targets, the red boxes are false positive ship targets, and green boxes are missed ship targets.}
	\label{fig:result ssdd}
\end{figure*}

\subsection{Comparison With No Interactions Between  Tasks}
The three tasks of the proposed method have positive interactions between each other. 
As shown in Fig.2, the detection module consists of CSPDarknet and a common block shared with the other two modules by equipping with residual blocks of feature extractor, instead of designing them separately.
DFF module integrates the denoising feature map with the original feature map by the proposed dual-feature fusion attention mechanism (descripted in Section II-B), which can not only eliminate the speckle noise, but also preserve the features of small targets.
Besides of sharing the same backbone to ensure the consistent of feature extraction, the segmentation task also employs the rotated Gaussian mask (descripted as Fig.6) based on the same ground-truth with object detection task.
Finally, WRBF is adopted to combine the predictions of object detection and the target segmentation, which is more conducive to the location of the target center point and the target direction.

In order to prove the effectiveness of the interactions between the three tasks, we implement experiments on  performing the three tasks in a sequential form, i.e, the denoising task, the segmentation task and the target detection task are executed separately as shown in Fig.\ref{fig:part}. 
Specifically, we firstly perform the denoising task with the L2 loss function to denoise SAR images. 
Then, we use the two independent backbone networks without sharing weights to extract features from the denoised images seperately, and perform target detection and segmentation tasks respectively without interaction between each other. 
Finally, the WRBF strategy is applied to fuse the detection and segmentation task results according the different confidence of target boxes.


\par 
The comparison results of the proposed MLDet and MLDet in sequential form are provided in Table \ref{fig: DIFFERENT NETWORK FORMS}, which indicates that the proposed MLDet is superior to  the sequential form in all indicators, including Precision, Recall, F1-score, $Ap_{50}$  and $Ap_{75}$.
Fig.\ref{fig:part-result} demonstrates some representative detection results by the proposed MLDet and MLDet in the sequential form.
On the premise of adopting the same WRBF strategy, the predicted orientation angle and center points of the bounding boxes by our proposed MLDet are more accurate than the results predicted in the sequential form, as shown from the first and the second visualizatin results in Fig.\ref{fig:part-result}. 
This is on account of the interactions between object detection module and segmentation module, including sharing the same of feature extraction layers, and deriving segmentations labels from the same ground-truth as the detection labels during training.
Besides that, the features of small objects are easily reduced during denoising by the denoised module in sequential form, which is unfavourable to the subsequent target detection and segmentation tasks. 
That is the reason why there are a large number of false and missed targets existed in the results of MLDet in sequential form, as shown in the third and the fourth  visualizatin results of Fig.\ref{fig:part-result}.

\begin{table}[h]
	\caption{ABLATION STUDIES OF \textbf{DIFFERENT NETWORK FORMS} ON SSDD+}
	\centering
	\setlength{\tabcolsep}{1.6mm}
	\begin{tabular}{lcccccc}
		\toprule[0.1mm]
		Methods        & Precision & Recall & F1-score  & $Ap_{50}$ &$Ap_{75}$ \\
		\midrule[0.1mm]
		MLDet in sequential form & 0.934	&	0.918	&	0.926 	&	0.928   &0.593 \\
		MLDet(Ours)    & \textbf{0.941}	&	\textbf{0.946}	&	\textbf{0.943}	&	\textbf{0.953}		&	\textbf{0.601}\\
		increment    & 0.7$\%$	& 2.8$\%$	&	1.7$\%$	&	2.5$\%$		&	0.8$\%$\\
		\bottomrule[0.1mm]
	\end{tabular}
	\label{fig: DIFFERENT NETWORK FORMS}
\end{table}

\subsection{Comparison with Other CNN-based Detection Methods}
\par To verify the generalization abilities and feasibility  of our  method, we implement experiments on two datasets of SSDD+ and HRSID. 
The recall rate, precision rate, F1-score, $AP_{50}$, $AP_{75}$ are employed to compare the performance with different methods, i.e., Faster R-CNN\cite{2017A35}, RetinaNet\cite{2017Focal15}, Oriented R-CNN\cite{2021Oriented}, CenterNet\cite{2019Objects25}, and YOLOX\cite{2021YOLOX}. 
The results in Table \ref{fig:data result all} indicate that our method MLDet outperforms other comparison methods mentioned above, achieving the state-of-the-art performance for both inshore and offshore ship detection. 
Fig. \ref{fig:PR} demenstrates  the precision-recall curves of ship detection results obtained by all of the compared methods. 
The purple curve in Fig. \ref{fig:PR} illustrates that our method is more precise than the other five CNN-based methods under different recall rates. An analysis of the experimental results is provided below.

\begin{figure*}[th]
	\centering
	\includegraphics[width=1\linewidth]{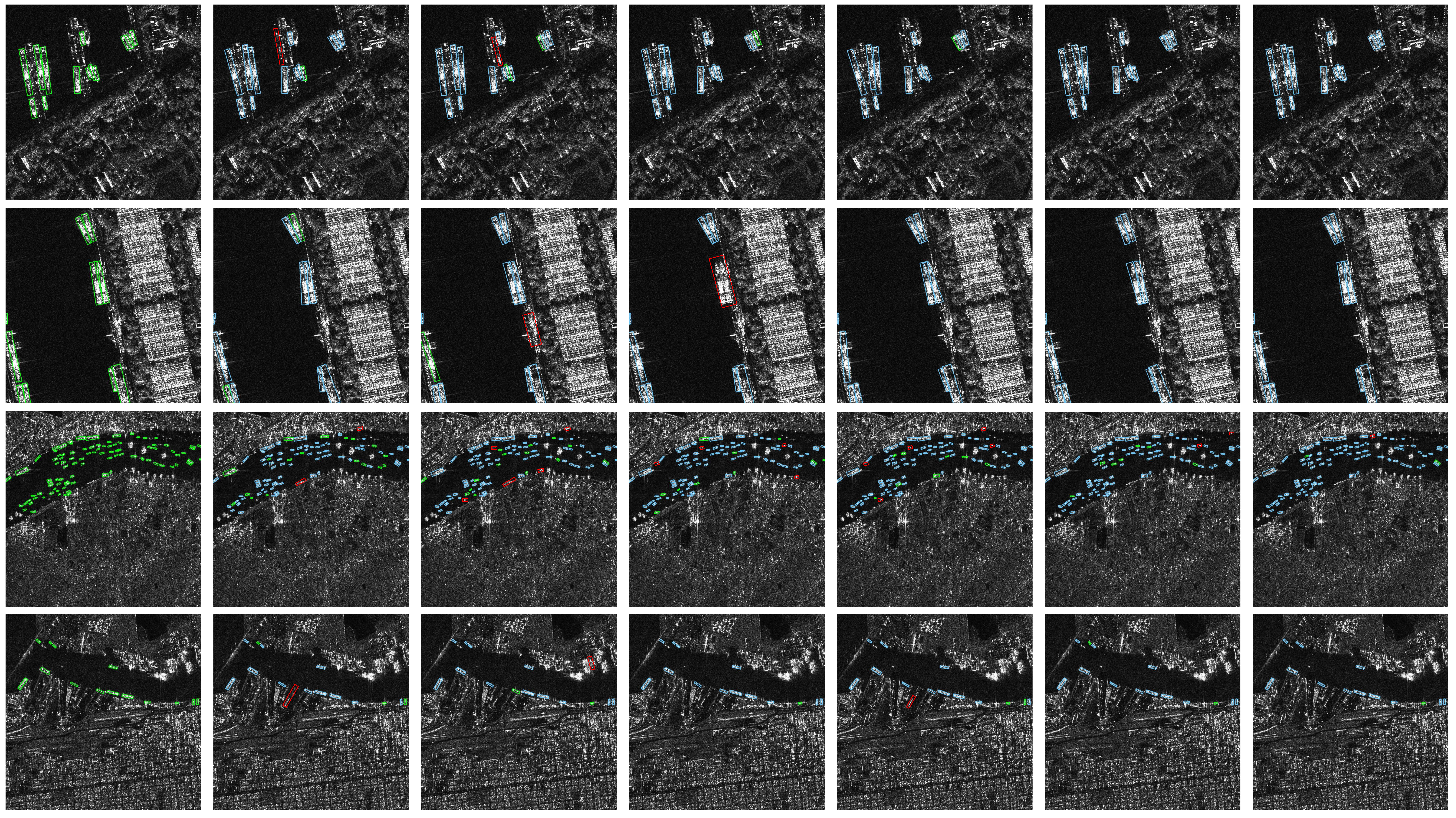}
	\begin{minipage}[t]{1\linewidth}      
		\vspace{0cm} 
		\qquad \qquad(a) \qquad \qquad \qquad (b) \qquad \qquad \quad (c)  \qquad  \qquad \qquad (d) \qquad \qquad \quad\quad (e) \qquad \qquad \quad\quad (f) \qquad \qquad \quad\quad (g)
	\end{minipage}

	\caption{Comparison of some detection results of different methods on HRSID. (a) The ground-truth. (b) Faster RCNN. (c) RetinaNet. (d) Oriented RCNN. (e) CenterNet. (f) YOLOX. (g) MLDet (our method). The first line shows  the detection results under complex background. The second line shows the detection results under  densely arranged ships. The third line shows the detection results  for small ships. The fourth line shows the detection results for multiscale  ships. Note that the blue boxes are true positive ship targets, the red boxes are false positive ship targets, and green boxes are missed ship targets.}
	\label{fig:result hrsid}
\end{figure*}

 \subsubsection{Comparison of Offshore Scene}
\par  As evidenced by our experimental results on SSDD+, our method is competitive and outperforms other classical methods for the detection of offshore vessels. 
As shown in Table \ref{fig:data result all}, our proposed method achieves improvements of 15.03$\%$, 8.74$\%$, 3.4$\%$,	4.61$\%$ and 2.86$\%$  over Faster R-CNN, RetinaNet, Oriented R-CNN, CenterNet and YOLOX on $Ap_{50}$, respectively, which benefits from the proposed DFF module and TS module. 
HRSID has a more complicated background and a large number of small ship objects in images, so that it is very suitable for evaluating the effectiveness of our proposed method for complex scenes.
Similarly, the proposed method outperforms other five compared methods, achieving the improvements of 11.1$\%$,	10.68$\%$,	7.45$\%$,	3.091$\%$	and	5.05$\%$  over Faster R-CNN, RetinaNet, Oriented R-CNN, CenterNet and YOLOX on $Ap_{50}$, respectively. 
The results in Table \ref{fig:data result all} indicate that our MLDet reaches the highest precision rate and recall rate under offshore scenes. 
As a result of extracting more characteristic features, the DFF module is robust to speckle noise. It concentrates more on significant features and establishes stronger links among context information features of ships than other methods. 
In addition, by capturing context information for multiscale objects, particularly small objects, the TS module alleviates the challenge of multiscale objects. 
In general, the proposed method   achieves the best performance compared with other existing network models.

\subsubsection{Comparison of Inshore Scene}
\par Using deep learning technology to detect a ship target in an inshore scene is more challenging than detecting a ship target in an offshore scene due to the interference caused by the inshore area. 
The detectors are very easily to incorrectly identify inferences in the scene of inshore  as ships as well as  causing the false alarms.  Under different model configurations, the detection accuracy of offshore scenes was found to be significantly higher than that of inshore scenes, as shown in Table  \ref{fig:data result all}.

Detection recall rates for offshore scenes are much higher than those for the scenes of inshore with different configurations of models. 
In order to evaluate a detector, it is important to consider the detection performance of inshore vessels. 
It can be observed from Table \ref{fig:data result all} that our method obtains an obvious improvement in the inshore scenes. Specifically, compared to the state-of-the-art methods, our method can  increase the  $Ap_{50}$ by ranging from $1.04\%$ to $11.2\%$ on SSDD+ and ranging from $1.84\%$ to $8.7\%$ on HRSID in the inshore scene, which is mainly because Use DFF modules can reduce the impact of complex backgrounds on ship object detection to some extent, resulting from the better detection performance of densely arranged ships. 
At the same time, the TS module  concentrates more on significant features and establishes stronger links among context information features of ships than other methods. 
Our method MLDet enhances the precision rate for the detection of inshore ships by a considerable margin, and also significantly reduces the number of false alarms. 
MLDet also achieves a much higher  $Ap_{75}$ than other methods, indicating that it is capable of achieving more accurate localization.

\subsubsection{Visual Results and Insight}
\par Fig. \ref{fig:result ssdd} and Fig. \ref{fig:result hrsid}  demonstrate some representative visualization results of the six compared methods in different cases, i.e., under complex background, small objects, densely arranged ships, and multiscale and small object detection. 
As shown in the first row and third row of Fig. \ref{fig:result ssdd} and Fig. \ref{fig:result hrsid}, the proposed method is superior to other methods  for  densely arranged ship detection under complex background.
It indicates intuitively that the TS module can achieve the accurate ship locations and recognitions in complex scenes.
Moreover, the proposed method is also effective for detecting  the densely arranged ships, as shown in the second row of Fig. \ref{fig:result ssdd} and Fig. \ref{fig:result hrsid}. 
Compared to the other five methods, our method does not produce any false positives or any false negatives. 
The fouth row in Fig. \ref{fig:result ssdd} and Fig. \ref{fig:result hrsid}  shows the detection results for multiscale and small objects. 
In the inshore area, there are many bright areas that are similar to ship objects, which is likely to cause severe false alarms. 
According to the comparisons, MLDet performs better than all other methods, especially for densely docked vessels. 
This validates that multitask learning is able to enhance the representation ability of our proposed network and effectively overcome false alarms in complex scenarios. By capturing multiscale context information, our method provides good detection results. 
The visualization results indicate that our method can effectively resolve the major problems associated with ship detection in SAR images.

\subsubsection{Comparison of Time and Model Complexity}
Table \ref{fig:data result all} presents the running time, parameter size, and model volumes of the six CNN-based methods verified. 
As presented in Table \ref{fig:data result all}, compared with the anchor free method CenterNet, there is a slight increase in both of the model parameters and the inference time per image by the proposed MLDet, which achieves remarkable detection accuracy. 
Region proposal causes the waste of storage and  computing resources, and also increases the computational complexity of the network models. 
Therefore, compared with two-stage methods such as Faster RCNN and Oriented R-CNN, MLDet has less model parameters and inference time. 
In spite of the fact that the proposed method does not inference as quickly as YOLOX, it can fully meet the requirements of real-time detection, and meanwhile  achieves remarkable accuracy of ship detections. 
Besides that, the detection results of the proposed MLDet on multiple datasets demonstrate its fine generalizability for different image data.

\begin{table*}
	\centering
	\setlength{\tabcolsep}{0.8mm}
	\caption{COMPARISON WITH OTHER DEEP LEARNING BASED METHODS ON \textbf{SSDD+} AND \textbf{HRSID}}
	\begin{tabular}{ll|cccccc|cccccc|cc}
		\toprule[0.3mm]
		\multirow{2}{*}{Dataset} & \multirow{2}{*}{Methods} & \multicolumn{6}{l|}{~ ~ ~ ~ ~ ~ ~ ~ ~ ~ ~ ~ ~ ~ ~ ~ ~ inshore} & \multicolumn{6}{l|}{~ ~ ~ ~ ~ ~ ~ ~ ~ ~ ~ ~ ~ ~ ~ ~ ~offshore} &
		\multicolumn{2}{l}{~ ~ ~ ~ ~ ~ ~ ~ ~ ~ ~ ~ ~ ~}  \\ 
		
		&                                 & Precision & Recall & F1-score & $Ap_{50}$  & $Ap_{75}$      &Runtime(ms)       & Precision & Recall & F1-score & $Ap_{50}$    & $Ap_{75}$ & Runtime(ms) & Params(M))& Volume(MB)\\ 
		\toprule[0.1mm]
		\multirow{6}{*}{SSDD+}  &	Faster R-CNN&	0.801 	&	0.880 	&	0.839 		&	0.823 &	0.530 & 21.4	&	0.902  	&	0.893 	&	0.897 	&	0.811 	&	0.502 &49.3	&58.5 &181.5\\
		&	RetinaNet	&	0.856 	&	0.816 	&	0.836 	&	0.849 	&	0.552 &11.3	&	0.794 	& 	0.848 	&	0.820 	&	0.874 	&	0.580&35.2 &48.1&103.6	\\
		&	Oriented R-CNN	&	0.900 	&	0.836 	&	0.867 	&	0.881 	&	0.591 &45.1	&	0.906 	&	0.895 	&	0.900 	&	0.927 	&	0.643 &74.3&63.5&206.4	\\
		&	CenterNet	&	0.943 	&	0.899 	&	0.920 	&	0.895 	&	0.613 &23.4	&	0.879 	&	0.893 	&	0.886 	&	0.915 	&	0.653 &52.7 & 	\textbf{25.8}& 	\textbf{83.3}	\\
		&	YOLOX	&	0.942 	&	0.898 	&	0.919 	&	0.925 	&	0.645 &	\textbf{10.3}	&	0.917 	&	0.922 	&	0.920 	&	0.932 	&	0.674 & 	\textbf{29.7} &50.4&130	\\
		&	MLDet (Ours)	&	\textbf{0.953} 	&	\textbf{0.945} 		&	\textbf{0.949} &		\textbf{0.935} &	\textbf{0.689} &14.2	&	\textbf{0.936} 	&	\textbf{0.948} 	&	\textbf{0.942} 	&	\textbf{0.961} 	&	\textbf{0.704} &30.4 &53.5 &152.8\\
		\toprule[0.1mm]
		\multirow{6}{*}{HRSID}     &	Faster	R-CNN	&	0.807 	&	0.831 	&	0.819 	&	0.814 	&	0.514 &34.1	&	0.832 	&	0.810 	&	0.821 	&	0.839 	&	0.533 &64.5	&58.5 &181.5\\
		
		&	RetinaNet	&	0.812 	&	0.821 	&	0.816 	&	0.824 	&	0.529 &20.1	&	0.832 	&	0.832 	&	0.832 	&	0.842 	&	0.561 &45.7 &48.1&103.	\\
		&	Oriented R-CNN	&	0.871 	&	0.796 	&	0.832 	&	0.830 	&	0.531 &60.4	&	0.858 	&	0.850 	&	0.854 	&	0.874 	&	0.581 &90.2 &63.5&206.4	\\
		&	CenterNet	&	0.846 	&	0.882 	&	0.864 	&	0.897 	&	0.695 &30.1	&	0.940 	&	0.852 	&	0.894 	&	0.918 	&	0.658 &69.6  & 	\textbf{25.8}& 	\textbf{83.3}	\\
		&	YOLOX	&	0.906 	&	0.873 	&	0.889 	&	0.883 	&	0.601 &	\textbf{15.3}	&	0.871 	&	0.858 	&	0.865 	&	0.898 	&	0.625 & 	\textbf{38.5} &50.4&130	\\
		&	MLDet (Ours) &	\textbf{0.911} 	&	\textbf{0.904} 	&	\textbf{0.907} 	&	\textbf{0.901} 	&	\textbf{0.667}	&19.4&	\textbf{0.927} 	&	\textbf{0.920} 	&	\textbf{0.924} 	&	\textbf{0.949} 	&	\textbf{0.695}	&42.9 &53.5 &152.8\\
		\bottomrule[0.3mm]
		\label{fig:data result all}
	\end{tabular}
\end{table*}

\section{CONCLUSION}
A novel multitask learning framework MLDet for SAR ship detection is proposed in this paper, which consists of a main detection module and two auxiliary task learning modules. Specifically, speckle supression and target segmentation are advantageous for enhancing the detection quality for SAR images. 
By embedding a denoising feature fusion module, the backbone is robust to speckle noise and focuses on feature extraction for ship targets.
A target segmentation module is proposed to help  the  network to extract the object regions of intersect from the cluttered background and improve the detection efficiency through pixel-by-pixel predictions. 
The ARWS loss function in object detection  can effectively avoid a sharp increase in the loss on the network training. Besides that, a WRBF strategy  is adopted to combine the predictions of rotated object detection and target segmentation, which improves the generalization ability of SAR ship detectors.
Experimental results on SSDD+ and HRSID datasets prove that the proposed method outperforms all the compared methods. 
As future work, we consider the method of parameter pruning and sharing to reduce redundant parameters insensitive to performance for convolution layer and full connection layer.

\end{document}